\makeatletter\def\graphicscache@inhibit{true}\makeatother

\documentclass[a4paper, 10pt, conference]{ieeeconf}      %

\IEEEoverridecommandlockouts                              %

\usepackage[style=ieee,hyperref,natbib=true,backend=bibtex,firstinits,doi=false,%
     mincitenames=1,maxcitenames=2,maxbibnames=99,sorting=none,terseinits=false,hyperref=true]{biblatex}

\bibliography{references.bib}

\usepackage{amsmath,amssymb,amsfonts}
\usepackage{pifont}
\usepackage{textcomp}
\usepackage{multirow}
\renewbibmacro*{bbx:savehash}{}%
\defbibheading{bibliography}[\bibname]{\section*{References}}

\usepackage{pgfplots}
\pgfplotsset{compat=newest}
\usepackage{tikz}

\usetikzlibrary{fit,arrows,arrows.meta,automata,backgrounds,calc,chains,%
decorations.markings,decorations.pathreplacing,decorations.pathmorphing,%
matrix,positioning,shapes,shapes.geometric,shapes.symbols,spy,trees,tikzmark,patterns}

\usepackage{hyperref}
\usepackage{url}
\usepackage{todonotes}
\usepackage{graphicx}
\usepackage{booktabs}
\usepackage{acronym}
\usepackage[capitalize]{cleveref}
\usepackage{siunitx}
\newcommand{\smartedge}{smart edge\ }

\usepackage{multirow}
\usepackage{gensymb}
\usepackage{algorithm}
\usepackage[beginComment=\#~, beginLComment=\#~, endLComment= ]{algpseudocodex}

\DeclareMathOperator{\sign}{sign}

\acrodef{FoV}{Field of View}
\acrodef{LiDAR}{Light Detection and Ranging}

\title{\LARGE \bf
Anticipating Human Behavior for Safe Navigation and\\Efficient Collaborative Manipulation with Mobile Service Robots 
}

\author{Simon Bultmann, Raphael Memmesheimer, Jan Nogga, Julian Hau, and Sven Behnke%
\thanks{This work was funded by grant BE 2556/16-2 (Research Unit FOR 2535 Anticipating Human Behavior) of the German Research Foundation (DFG)}%
\thanks{All authors are with the Autonomous Intelligent Systems group, %
		University of Bonn, Germany; {\tt\scriptsize\{bultmann|memmesheimer\}@ais.uni-bonn.de}}%
}

\begin{document}
\maketitle
\thispagestyle{empty}
\pagestyle{empty}

\begin{abstract}
The anticipation of human behavior is a crucial capability for robots to interact with humans safely and efficiently.
We employ a smart edge sensor network to provide global observations, future predictions, and goal information to integrate anticipatory behavior for the control of a mobile manipulation robot.
We present approaches to anticipate human behavior in the context of safe navigation and collaborative mobile manipulation.
First, we anticipate human motion by employing projections of predicted human trajectories from smart edge sensor observations into the planning map of a mobile robot.
Second, we anticipate human intentions in a collaborative furniture-carrying task to achieve a given room layout.
Our experiments indicate that anticipating human behavior allows for safer navigation and more efficient collaboration.
Finally, we showcase an integrated robotic system that anticipates human behavior while collaborating with an operator to achieve a target room layout, including the placement of tables and chairs.
\end{abstract}

\section{Introduction}
Humans anticipate the actions of others in their surroundings and plan their own actions accordingly. 
This ability makes interaction more intuitive, efficient, safe, and natural.
Robots, however, often lack this ability and are perceived as unpredictable, leading to unsafe interactions~\cite{hoffman2010anticipation,huang2016anticipatory}.

While much research has been conducted on anticipating human behavior~\cite{hoffman2010anticipation,psarakis2022fostering,canuto2021action,duarte2018action}, we found that the actual integration of anticipatory behavior into mobile manipulation robot planning remains underrepresented.
This work presents approaches for anticipating human behavior in two natural interaction and collaboration scenarios.

First, we anticipate human behavior by encoding future human trajectories observed by an allocentric \smartedge sensor network into the planning map of an autonomously navigating robot.
This allows the robot to incorporate globally observed predictions into its local planning system and navigate more safely among humans.
The \smartedge sensor system enables the incorporation of predictions that cannot be anticipated with local sensory inputs alone.

Second, we anticipate human behavior in the context of collaborative furniture handling.
In this scenario, a mobile manipulation platform collaborates with a human to move tables and chairs to achieve a predefined room layout.
Anticipation in this context combines compliant collaborative control with goal anticipation given the target layout.

Prior existing work considered robots as nodes in a \smartedge sensor network~\cite{bultmann2023external} for collaborative semantic mapping.
We build on this, but now focus on anticipating human behavior based on the allocentric 3D semantic scene model built from smart edge sensor observations and making the robot act on these anticipations.

\begin{figure}[t]
  \centering
  \begin{tikzpicture}
  \node[anchor=north west,inner sep=0] (image) at (0, 0){\includegraphics[height=3.63cm]{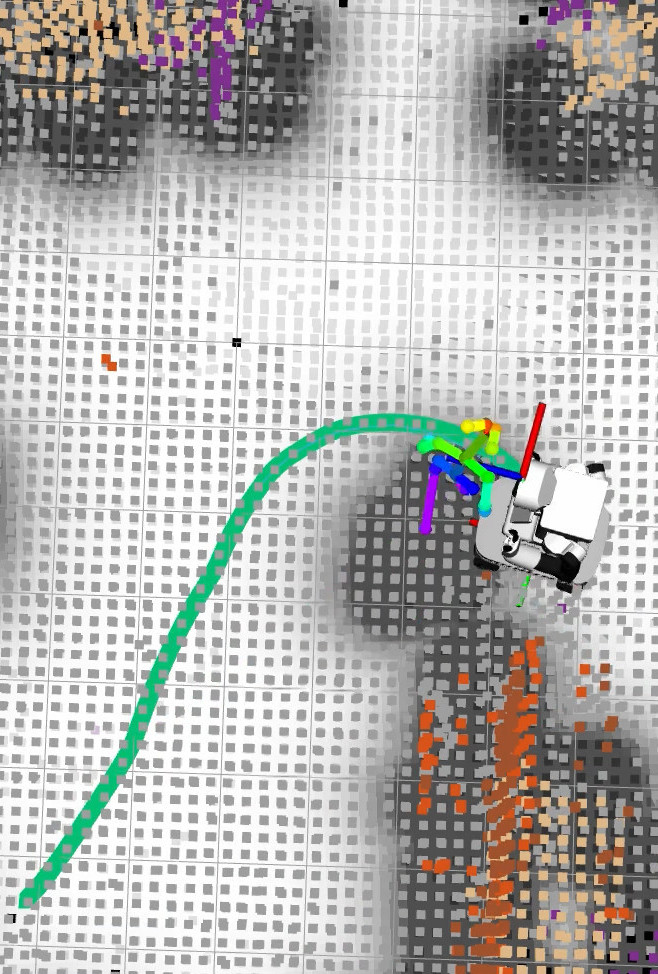}};
  \node[anchor=north west,inner sep=0, xshift=0.2em] (image1) at (image.north east){\includegraphics[height=3.63cm]{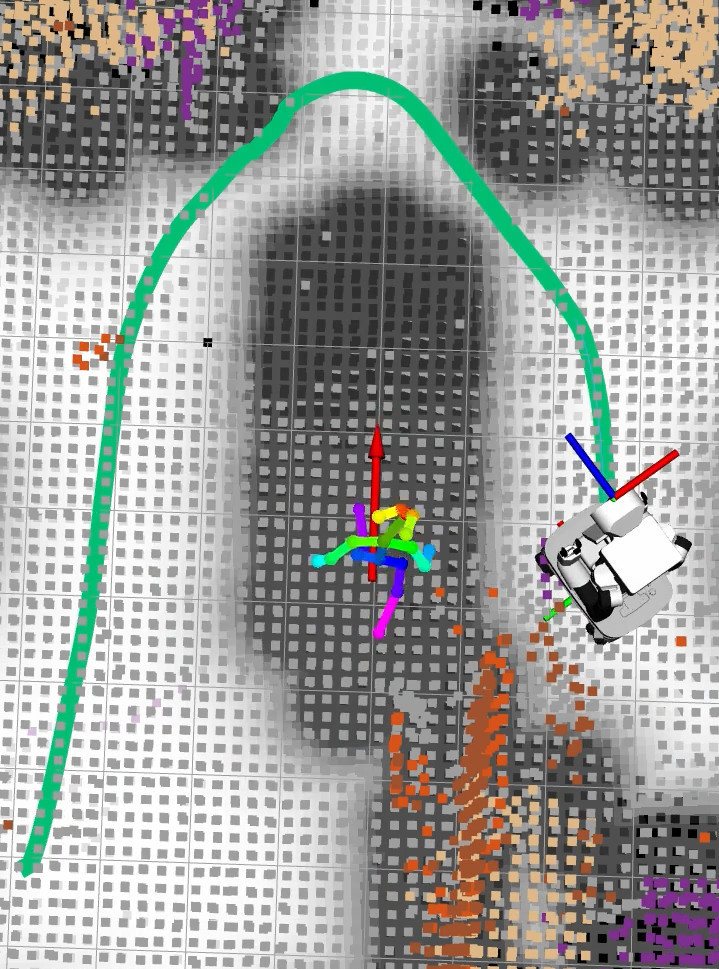}};
  \node[anchor=north west,inner sep=0, xshift=0.5em] (image2) at (image1.north east){\includegraphics[width=3.cm, trim=0 25 0 0, clip]{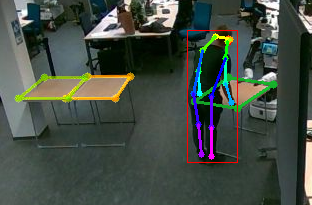}};
  \node[anchor=north west,inner sep=0, yshift=-0.2em] (image3) at (image2.south west){\includegraphics[width=3.cm]{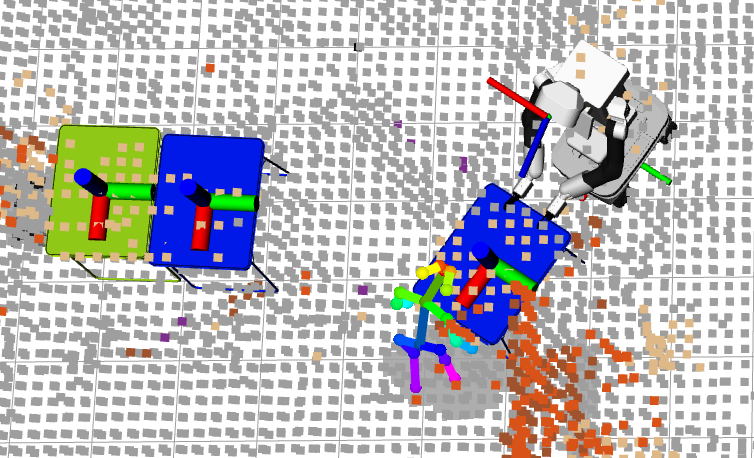}};
  \node[inner sep=0.15em,scale=.5, anchor=south west,yshift=0.em, rectangle, align=left, fill=lightgray, font=\sffamily] (n_0) at (image.south west) {3D scene view};
  \node[inner sep=0.15em,scale=.5, anchor=south west,yshift=0.em, rectangle, align=left, fill=lightgray, font=\sffamily] (n_0) at (image1.south west) {3D scene view};
  \node[inner sep=0.15em,scale=.5, anchor=south west,yshift=0.em, rectangle, align=left, fill=lightgray, font=\sffamily] (n_0) at (image2.south west) {External camera view};
  \node[inner sep=0.15em,scale=.5, anchor=south west,yshift=0.em, rectangle, align=left, fill=lightgray, font=\sffamily] (n_0) at (image3.south west) {3D scene view};
  \node[inner sep=0.,scale=.9, anchor=north west,yshift=-0.3em, xshift=.7em, rectangle, align=left, font=\scriptsize\sffamily] (n_0) at (image.south west) {(a) navigation without (l) and with (r) anticipation:\\path is adapted before robot can see the person};
  \node[inner sep=0.,scale=.9, anchor=north,yshift=-0.3em, rectangle, align=left, font=\scriptsize\sffamily] (n_0) at (image3.south) {(b) collaborative manipulation:\\where to grasp and place table};
  
  \end{tikzpicture}
  \vspace{-.7em}
  \caption{Two scenarios in which our robot anticipates human behavior.}
  \label{fig:teaser}
  \vspace{-1.em}
\end{figure}

\cref{fig:teaser} illustrates the two scenarios in which we anticipate human behavior.
For the navigation scenario, we anticipate future human trajectories and adapt the robot's navigation path accordingly.
For the collaborative manipulation scenario, we anticipate the human's intention to decide from where to grasp which table. During carrying, we incorporate the anticipated target pose for collaborative manipulation.

\noindent The main contributions of this paper are as follows:
\begin{itemize}
  \item We propose an approach for anticipating human motion for safe, human-aware navigation.
  \item We present an approach for anticipating collaborative human behavior in the context of furniture carrying to achieve a given target configuration.
  \item We conduct real-robot experiments with two subjects to evaluate the performance of our proposed methods.
  \item We successfully demonstrate the benefits of anticipating human behavior in the context of safe navigation and collaborative manipulation with a mobile service robot.
\end{itemize}

\begin{figure*}[t]
	\centering
	\resizebox{0.95\linewidth}{!}{%
\begin{tikzpicture}[font=\footnotesize\sffamily,on grid,>={Stealth[inset=0pt,length=4pt,angle'=45]}]
		\tikzset{every node/.append style={node distance=3.0cm}}
		\tikzset{img_node/.append style={minimum size=1.5em,minimum height=3em,align=center}}
		\tikzset{content_node/.append style={scale=0.85,minimum size=1.5em,minimum height=2em,minimum width={width("\& Anticipation") + 0.7em},draw,align=center,fill=blue!15!white}, rounded corners}
		\tikzset{content_node2/.append style={scale=0.85,minimum size=1.5em,minimum height=2em,minimum width={width("Cameras") + 0.7em},draw,align=center,fill=blue!15!white}, rounded corners}
		\tikzset{content_node3/.append style={scale=0.85,minimum size=1.5em,minimum height=2em,minimum width={width("Manipulation") + 1.em},draw,align=center,fill=blue!15!white}, rounded corners}
		\tikzset{label_node/.append style={scale=0.8, near start}}
		\tikzset{junction/.append style={circle, fill=black, minimum size=3pt, draw}}

		\definecolor{red}{rgb}     {0.9,0.0,0.0}
		\definecolor{green}{rgb}   {0.0,0.5,0.0}
		\definecolor{blue}{rgb}    {0.0,0.0,0.5}
		\definecolor{grey}{rgb}    {0.5,0.5,0.5}
		\definecolor{light_grey}{HTML}  {CCCCCC}
		
		\node(rgbd1)[content_node2,fill=green!15!white,anchor=north west,minimum height=3em] at (0, 0) {RGB-D\\camera};
		\node(dets)[content_node2,anchor=west, xshift=0.5cm] at (rgbd1.east) {\setlength{\tabcolsep}{3pt}\begin{tabular}{r c} Person- & \multirow{ 3}{*}{\shortstack{Detections\\\& keypoints}}\\ Robot- \\ Object- \end{tabular}};
		\draw[dotted, very thick] (dets.east) ++(1em, 0.em) -- ++(.9em, 0.em);
		
		\node(pt1)[xshift=.1em, yshift=-.1em] at (rgbd1.west |- dets.north){};
		\node(pt2)[xshift=-.1em, yshift=-.1em] at (dets.north east){};
		\node(pt3)[xshift=-.1em, yshift=.1em] at (dets.south east){};
		\node(pt4)[xshift=.1em, yshift=.1em] at (rgbd1.west |- dets.south){};
		\draw[thick, rounded corners, grey!60!white] (pt1.north west) -- (pt2.north east)-- (pt3.south east) -- (pt4.south west) -- cycle;
		
		\node(pt11)[xshift=-.2em, yshift=.2em] at (rgbd1.west |- dets.north){};
		\node(pt21)[xshift=3.5em, yshift=.2em] at (dets.north east){};
		\node(pt31)[xshift=3.5em, yshift=-.2em] at (dets.south east){};
		\node(pt41)[xshift=-.2em, yshift=-.2em] at (rgbd1.west |- dets.south){};
		\begin{scope}[on background layer]
		\draw[thick, rounded corners, grey!20!white,fill] (pt11.north west) -- (pt21.north east)-- (pt31.south east) -- (pt41.south west) -- cycle;
		\end{scope}
		
		\node(label_sensors)[scale=0.9,anchor=south west,xshift=-0.2cm,yshift=0.2cm] at (rgbd1.north west) {\textbf{Smart edge sensors $1,\,\ldots,\,N$}};
		\draw[->, thick] (rgbd1.east) -- (dets.west);

		\node(rgbd2)[content_node2,fill=green!15!white,anchor=north west, yshift=-1.0cm] at (rgbd1.south west) {RGB-D};
		\node(robot_percept)[content_node3,anchor=west, xshift=0.3cm] at (rgbd2.east) {Perception \&\\manipulation};
		
		\node(robot_nav)[content_node3,anchor=north west, yshift=-0.1cm] at (robot_percept.south west) {Navigation};
		\node(lidar)[content_node2,fill=green!15!white,anchor=east, xshift=-0.3cm] at (robot_nav.west) {LiDAR};
		\node(context)[content_node3,minimum height=4.5em,anchor=north west, xshift=0.3cm, draw=red, thick] at (robot_percept.north east) {Incorporation\\of global\\context};
		
		\node(pt12)[xshift=-.2em, yshift=.1em] at (rgbd1.west |- context.north){};
		\node(pt22)[yshift=.1em] at (pt21 |- context.north){};
		\node(pt32)[yshift=-.1em] at (pt31 |- context.south){};
		\node(pt42)[xshift=-.2em, yshift=-.1em] at (rgbd1.west |- context.south){};
		\begin{scope}[on background layer]
		\draw[thick, rounded corners, grey!20!white,fill] (pt12.north west) -- (pt22.north east)-- (pt32.south east) -- (pt42.south west) -- cycle;
		\end{scope}
		
		\node(label_robot)[scale=0.9,anchor=south west,xshift=-0.2cm,yshift=0.15cm] at (rgbd2.north west) {\textbf{TIAGo++ robot}};
		\draw[->, thick] (rgbd2.east) -- (robot_percept.west);
		\draw[->, thick] (lidar.east) -- (robot_nav.west);
		
		\draw[->, thick, red] (robot_percept.east -| context.west) -- (robot_percept.east);
		\draw[->, thick, red] (robot_nav.east -| context.west) -- (robot_nav.east);

		\node(mvfusion)[content_node,anchor=north west, xshift=5cm, yshift=0.1cm] at (dets.north east) {Multi-view\\fusion};
		\node(poseest)[content_node,anchor=north west, yshift=-0.2cm] at (mvfusion.south west) {3D pose\\estimation};
		\node(semmap)[content_node,anchor=north west, yshift=-0.2cm] at (poseest.south west) {Semantic\\ mapping};
		\node(control)[content_node,anchor=north west, yshift=-0.2cm, draw=red, thick] at (semmap.south west) {Task control\\\& anticipation};
		
		\node(pt13)[xshift=.1em, yshift=-.1em] at (mvfusion.north west){};
		\node(pt23)[xshift=-.1em, yshift=-.1em] at (mvfusion.north east){};
		\node(pt33)[xshift=-.1em, yshift=.1em] at (control.south east){};
		\node(pt43)[xshift=.1em, yshift=.1em] at (control.south west){};
		\begin{scope}[on background layer]
		\draw[thick, rounded corners, grey!20!white,fill] (pt13.north west) -- (pt23.north east)-- (pt33.south east) -- (pt43.south west) -- cycle;
		\end{scope}
		
		\node(ptl1)[xshift=3.8em] at (dets.20 -| dets.east){};
		\node(ptl2)[xshift=-.2em] at (dets.20 -| mvfusion.west){};
		\node(ptl3)[xshift=3.8em] at (dets.350 -| dets.east){};
		\node(ptl4)[xshift=-.2em] at (dets.350 -| mvfusion.west){};
		\draw[->, thick] (ptl1) -- node[label_node,midway,above] {Semantic percepts} (ptl2);
		\draw[->, thick, red] (ptl4) -- node[label_node,midway,above] {Semantic feedback} (ptl3);
		
		\node(ptl11)[xshift=0em] at (context.40 -| context.east){};
		\node(ptl21)[xshift=-.2em] at (context.40 -| mvfusion.west){};
		\node(ptl31)[xshift=0em] at (context.348){};
		\node(ptl41)[xshift=-.2em] at (context.348 -| mvfusion.west){};
		\draw[->, thick] (ptl11) -- node[label_node,near start,above, text width=10em,align=right] {Robot percepts\\\& odometry} node[label_node,midway,below] {Task progress}(ptl21);
		\draw[->, thick, red] (ptl41) -- node[label_node,midway,above] {Semantic feedback} node[label_node,midway,below, xshift=1.6em, text width=10em,align=left] {Localization,\\obstacles, intentions} (ptl31);
		
		\node(label_backend)[scale=0.9,anchor=south west,xshift=-0.2cm,yshift=0.1cm] at (mvfusion.north west) {\textbf{Backend}};
		
		\node(scene)[img_node,anchor=west, xshift=.7cm, yshift = -.45cm] at (poseest.east) {\setlength{\fboxsep}{0pt}\setlength{\fboxrule}{0.5pt}\fbox{\includegraphics[height=3.5cm]{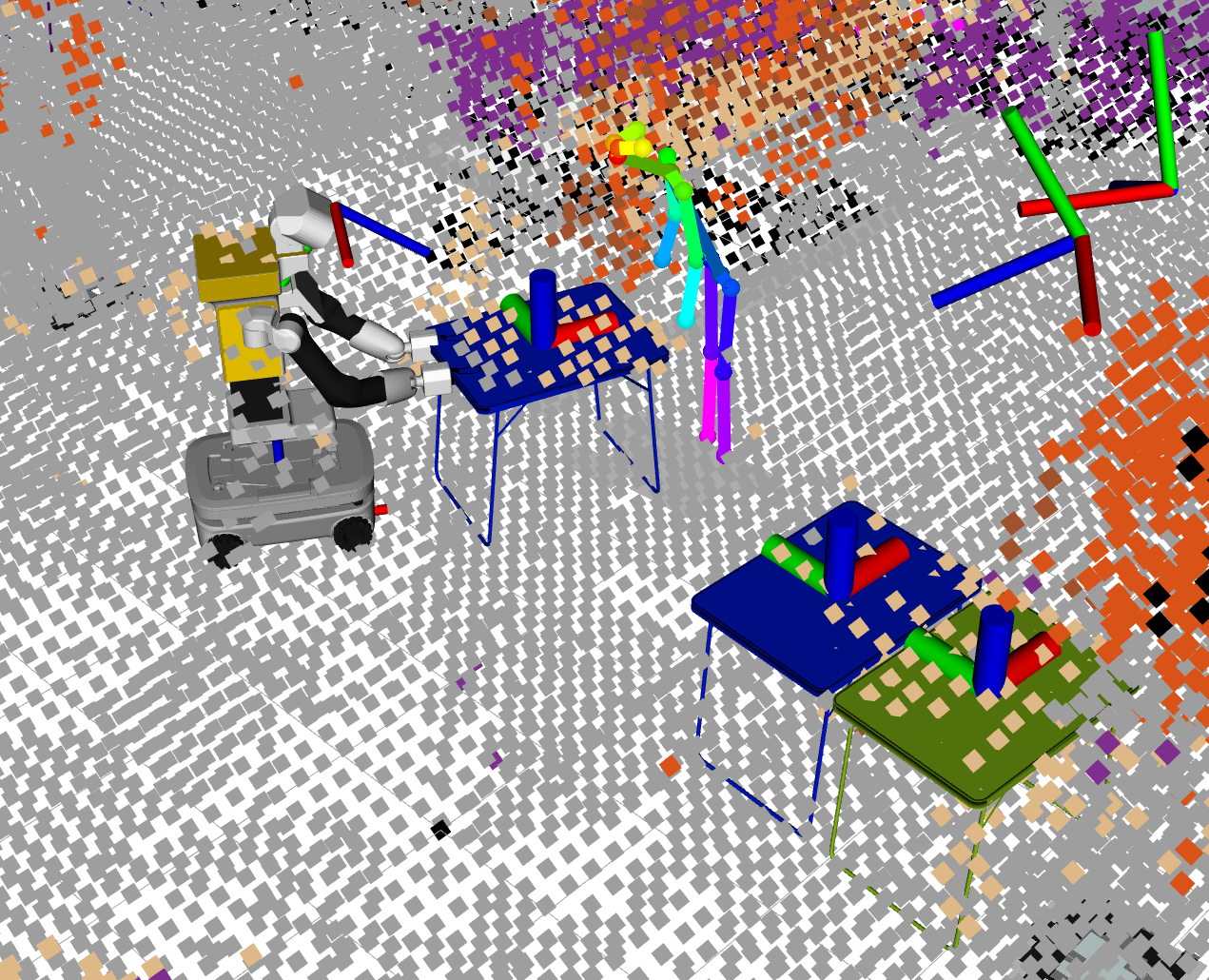}}};
		\node(label_scene)[scale=0.9,anchor=south west,xshift=0.1cm,yshift=-.2cm] at (scene.north west) {\textbf{Semantic scene model}};
		
		\draw[->, thick] (poseest.east |- scene.west) ++ (.35em, 0.) -- (scene.west);

\end{tikzpicture} 
 	}
	\vspace{-.7em}
\caption{Data processing architecture of the developed approach. A network of smart edge sensors supervises the work space. It detects persons, robots, and objects and estimates their pose. The backend fuses local percepts and controls the task. Both the robot and the sensors incorporate semantic feedback.}
	\label{fig:method}
	\vspace{-.7em}
\end{figure*}

\section{Related Work}

\paragraph*{\hspace*{-4.5ex}$\circ$\,Anticipation}
\citet{canuto2021action} analyze the influence of additional contextual cues (gaze, movement, object information) to improve action anticipation in a collaborative setting. The problem is formulated as a classification task to predict the next most likely action. 
\citet{duarte2018action} conducted experiments within a similar setup where humans had to anticipate the actions of a robot.
\citet{tanke2023social} anticipate future human movements using a diffusion model for future prediction.
The loss function is defined to emphasize the importance of social context.
In contrast to the above approaches, which focus on the analysis of anticipation, there is much less work on the actual implementation of human anticipation in a robotic system.
\citet{huang2016anticipatory} present a system that proactively performs actions to assist a human in a collaborative task based on the user's gaze patterns. Like our approaches, they observed an improved task completion time.
\citet{psarakis2022fostering} study human-robot collaboration in an industrial setting, focusing on the effects of robot anticipatory cues and adaptability on task efficiency, safety, and fluency.
\citet{barmann2023incremental} present an approach for dialog-based incremental learning through natural interaction. Large language models are employed for high-level behavior generation that adapts to future tasks.

\paragraph*{\hspace*{-4.5ex}$\circ$\,Human-Aware Navigation}
de~\citet{deheuvel23iros} adapt robot navigation to user preferences that have been explicitly demonstrated in a virtual-reality environment beforehand. The approach is evaluated in simulation only.
\citet{luber2012socially} present a learning-based approach for social-aware robot navigation. Criteria like path length and travel time, as well as objective criteria such as social comfort, are addressed by an unsupervised learning approach.
\citet{bruckschen2020human} present a foresighted navigation approach, ensuring social comfort while anticipating human trajectories at a target location to offer timely assistance. In contrast to our approach, they rely on the onboard sensors of the robot, while our approach integrates predictions by a \smartedge sensor network that allows for incorporating predictions that are not tractable by relying solely on the robot's observations.
Other approaches employ social force models to navigate among humans~\cite{helbing1995social,ferrer2017robot}.
\citet{arena2009learning} learn to navigate and anticipate actions in an environment based on sensory input. The anticipation of human behavior is not considered.

\paragraph*{\hspace*{-4.5ex}$\circ$\,Furniture Handling}
\citet{rus1995moving} were among the first to present an approach for collaborative mobile furniture movement. 
In their setting, in modern terms, simplistic robots were moving furniture by executing push sequences and aiming to enforce contact with furniture.
\citet{knepper2013ikeabot} presented an approach for coordinated multi-robot furniture assembly.  
A geometric specification is provided to their approach to generate a symbolic plan for the assembly of parts. 
The assembly plan is then sequentially executed by two collaborating robots equipped with special tool end-effectors. 
The vision part is bypassed by using motion capture systems. 
The approach has been exemplarily demonstrated for the assembly of an Ikea Lack table.  
A quantitative evaluation of the system has not been conducted.
Closely related to our proposed approach is the work of \citet{StucklerB:Humanoids11, StucklerSB:Frontiers16}.
They presented a compliant system for various manipulation tasks, including compliance control of human collaborative table carrying.
Upon detecting a person, the robot proceeds to approach and grasp the table, subsequently waiting for the individual to initiate lifting.
When the person lowers the table, the robot initiates a placement motion.
In contrast to our method, the robot does not actively anticipate the intended outcome.
The evaluation of the system is limited to a qualitative demonstration of the system in the context of the RoboCup@Home competition.
\citet{fallatah2021towards} present an approach for furniture arrangement of chairs.
A multipurpose screen setup is used to set the arrangement.
An overhead camera is employed to visualize the scenes on the screen and provides feedback to the robot.
The chairs are attached directly to the robots, facilitating their manipulation, and markers are attached to the chairs, simplifying the perception.
In contrast to the previous approaches, a user study was conducted to evaluate whether user expectations were met.

\section{Method}
\subsection{Overview}
An overview of the data processing architecture of our proposed approach is given in \cref{fig:method}.
A network of 25 smart edge sensors (cf. \cref{fig:setup}~(b)) is installed in our lab and gathers semantic observations of the scene, i.e., detections and keypoints of persons, robots, and objects, as well as semantic point clouds~\cite{Bultmann_RSS_2021,bultmann_ias2022,hau_objectlevel_2022}. The sensor views are fused on a central backend into an allocentric 3D semantic scene model $\mathcal{M}$ comprising a volumetric semantic map $\mathcal{V}$ of the static environment as well as dynamic human, robot, and object models $\mathcal{P}, \mathcal{R}, \mathcal{O}$, where:
\begin{align}
\mathcal{P} &= \left\{\left(\left\{\mathbf{kp}\right\}_{i=1}^{K} \in \mathbb{R}^3, \mathbf{v} \in \mathbb{R}^3 \right)\right\}\,, \label{eq:pers}\\
\mathcal{R} &= \left(\mathbf{M}_\text{r}, \mathbf{p}_\text{r} = \left[x_\text{r}, y_\text{r}, \theta_\text{r}\right]^\top\in \mathbb{SE}(2)\right. ,\label{eq:rob}\\
&\quad\quad\left.\mathbf{v}_\text{r} = \left[v_\text{x}, v_\text{y}, \omega_\text{$\theta$}\right]^\top \in \mathbb{R}^3 \right)\nonumber\,,\\
\mathcal{O} &= \left\{\left(\mathbf{M}_\text{o}, c, \mathbf{p}_\text{o} = \left[x_\text{o}, y_\text{o}, \theta_\text{o}\right]^\top\in \mathbb{SE}(2)\right)\right\}\label{eq:obj}\,.
\end{align}
The tracked persons $\mathcal{P}$ are represented by a set of $K$ keypoints that form the body skeleton and the root joint velocity. The robot $\mathcal{R}$ is represented by its mesh model $\mathbf{M}_\text{r}$, pose $\mathbf{p}_\text{r}$, and velocity $\mathbf{v}_\text{r}$ on the ground plane. We assume a single robot visible in the scene.
The objects $\mathcal{O}$ are represented by their mesh model $\mathbf{M}_\text{o}$, semantic class $c$, and pose $\mathbf{p}_\text{o}$ on the ground plane. We consider tables and chairs as object classes in our model.

The smart edge sensors receive parts of the semantic scene model $\mathcal{M}$ as semantic feedback to incorporate global context, e.g., about occlusions, into their local perception~\cite{Bultmann_RSS_2021,bultmann_ias2022}.
The robot, similar to the smart edge sensors, augments its local perception, manipulation, and navigation capabilities with global context information received as semantic feedback from the semantic scene model $\mathcal{M}$.

The robot receives feedback about (i)~its pose in the scene model,
(ii)~persons who are in its vicinity but are out of sight of its internal sensors, e.g., due to occlusions or limited \ac{FoV}, and their predicted movement,
as well as (iii)~the objects and the intended target configuration of the manipulation task.
This enables anticipatory human-aware robot navigation where the robot preemptively adjusts its navigation path, e.g., to persons appearing from behind occluders (cf. Algorithm~\ref{alg:nav}) or to reach the intended target pose for picking up or placing an object (cf. Algorithms~\ref{alg:approach},~\ref{alg:followhand}).

We detail the developed approaches for perception, navigation, and manipulation in the following.
\begin{figure}[t]
  \centering
  \begin{tikzpicture}
  \node[anchor=north west,inner sep=0] (image) at (0,0){\includegraphics[height=2.5cm, trim=175 14 195 110, clip]{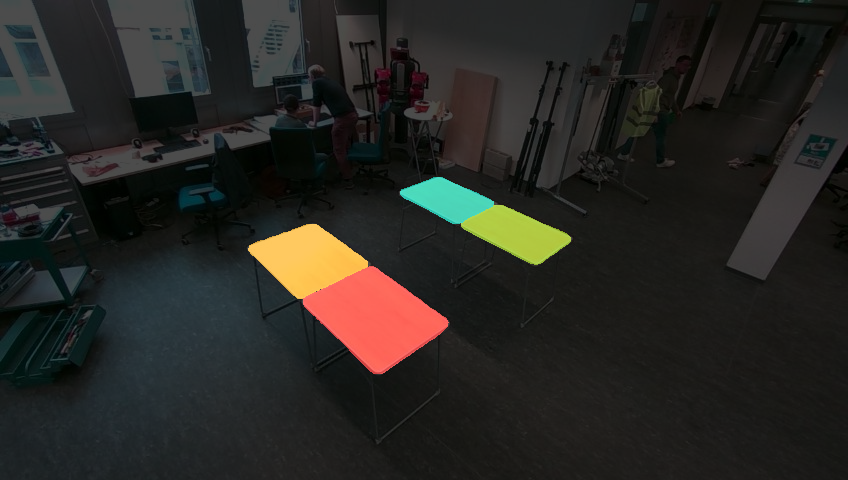}};
  \node[anchor=north west,inner sep=0, xshift=0.2em] (image1) at (image.north east){\includegraphics[height=2.5cm, trim=175 14 195 110, clip]{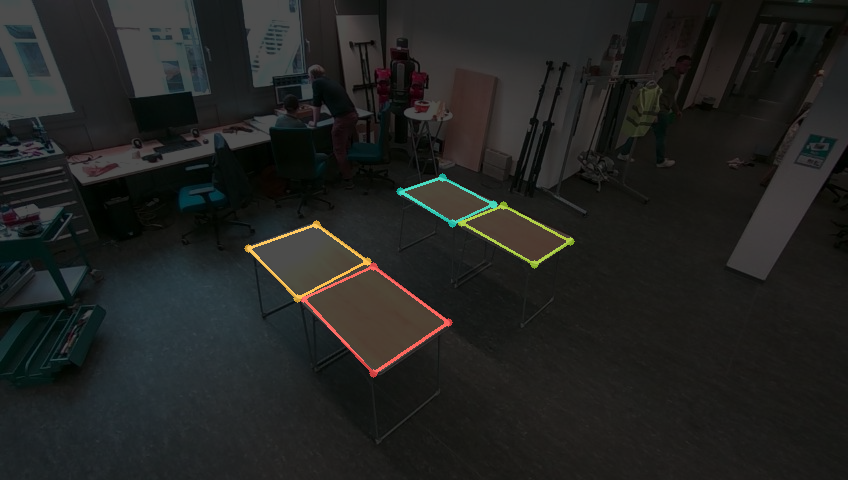}};
  \node[anchor=north west,inner sep=0, xshift=0.2em] (image2) at (image1.north east){\includegraphics[height=2.5cm, trim=175 14 195 110, clip]{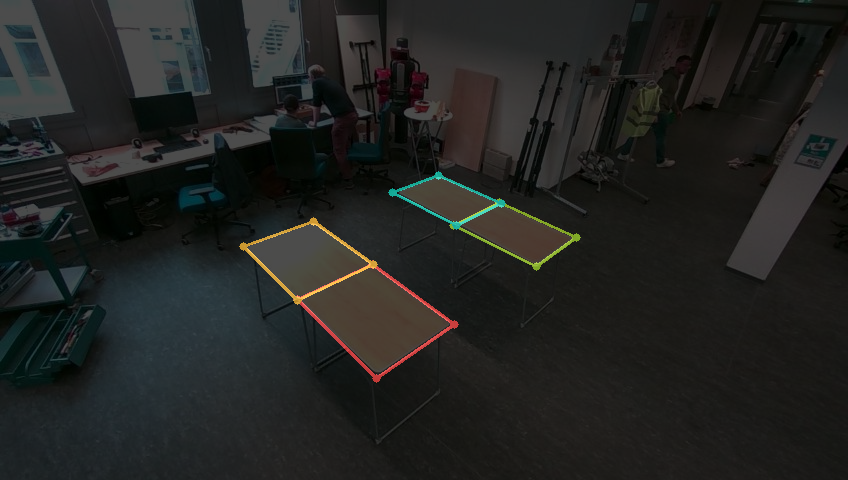}};
  
  \node[inner sep=0.,scale=.9, anchor=north,yshift=-0.3em, rectangle, align=left, font=\scriptsize\sffamily] (n_0) at (image.south) {(a) instance masks};
  \node[inner sep=0.,scale=.9, anchor=north,yshift=-0.3em, rectangle, align=left, font=\scriptsize\sffamily] (n_0) at (image1.south) {(b) initial keypoints};
  \node[inner sep=0.,scale=.9, anchor=north,yshift=-0.3em, rectangle, align=left, font=\scriptsize\sffamily] (n_0) at (image2.south) {(c) refined keypoints};
  \end{tikzpicture}
  \vspace{-1.8em}
  \caption{Table instance segmentation from smart edge sensor view, extracted contours, and keypoints.}
  \label{fig:instanceseg}
  \vspace{-1.3em}
\end{figure}

\subsection{Semantic Perception}
\label{sec:perception}
\paragraph*{\hspace*{-4.5ex}$\circ$\,Smart Edge Sensor Network}
The smart edge sensor boards gather semantic observations from multiple views, covering the entire scene from different directions. Thus, a complete, allocentric 3D semantic scene model $\mathcal{M}$ is estimated and updated over time, based on %
~\cite{Bultmann_RSS_2021,bultmann_ias2022,hau_objectlevel_2022}, comprising a semantic map of the static environment $\mathcal{V}$ as well as dynamic human, robot, and object poses $\mathcal{P}, \mathcal{R}, \mathcal{O}$ (Eqns.~\eqref{eq:pers}-\eqref{eq:obj}).

For robot localization by the external smart edge sensors, we adapt prior work~\cite{bultmann2023external}, developed for the Toyota HSR robot, to the employed TIAGo++ robot. While in~\cite{bultmann2023external}, only static keypoints on the rigid body were used for robot pose estimation, here, we also employ dynamic keypoints on the articulated robot body and dynamically adapt the reference points for pose estimation based on the robot joint configuration. The pose $\mathbf{p}_\text{r}$ from the robot model $\mathcal{R}$ (Eq.~\eqref{eq:rob}) estimated by the sensor network is sent to the robot to initialize and correct its localization in the scene.

We further extend the semantic perception to furniture objects (i.e., tables, chairs), with a focus on separating object instances that stand closely together side-by-side.

To separate individual objects in the local sensor views, we employ an instance segmentation model based on YOLO\-v8~\cite{Jocher_Ultralytics_YOLO_2023}, fine-tuned on synthetic data generated with Omniverse Replicator~\cite{nvidia_omniverse_replicator}.
For the table class, the model is trained to segment only the tabletop, allowing keypoints to be directly extracted from the instance masks (\cref{fig:instanceseg}~(a)).
Initial table keypoints are obtained from the mask contours with the Douglas-Peucker algorithm~\cite{douglas1973algorithms}, selecting four points that maximize the Intersection over Union (IoU) with the mask (\cref{fig:instanceseg}~(b)).
The resulting quadrilateral is refined by aligning its edges with the mask contour to precisely determine the four corner keypoints of each tabletop (\cref{fig:instanceseg}~(c)).
Object poses are estimated from 2D-3D keypoint correspondences via PnP~\cite{lepetit2009ep, collins2014infinitesimal}, restricted to 3\,DoF on the ground plane.
The table yaw angle $\theta_\text{t}$ is further restrained to $[0\ldots\pi]$ to account for their 180\degree~symmetry.
After removing outlier pose estimates, ICP refinement is performed between the mesh and the instance point cloud.
Chair keypoints and poses are computed following~\cite{hau_objectlevel_2022}, using chair crops derived from the instance segmentation.
The resulting object poses are transmitted to the central backend, where they are fused with other sensor views and tracked using a Kalman filter~\cite{kalman1960new}.

The obtained object models $\mathcal{O}$ (Eq.~\eqref{eq:obj}) are used together with the tracked persons $\mathcal{P}$ (Eq.~\eqref{eq:pers})~\cite{Bultmann_RSS_2021} to anticipate the robot target poses for pickup and placement of the furniture objects (cf. Sec.~\ref{sec:nav}).

\begin{figure}[t]
  \centering
  \begin{tikzpicture}
  \node[anchor=north west,inner sep=0] (image) at (0,0){\includegraphics[height=2.5cm]{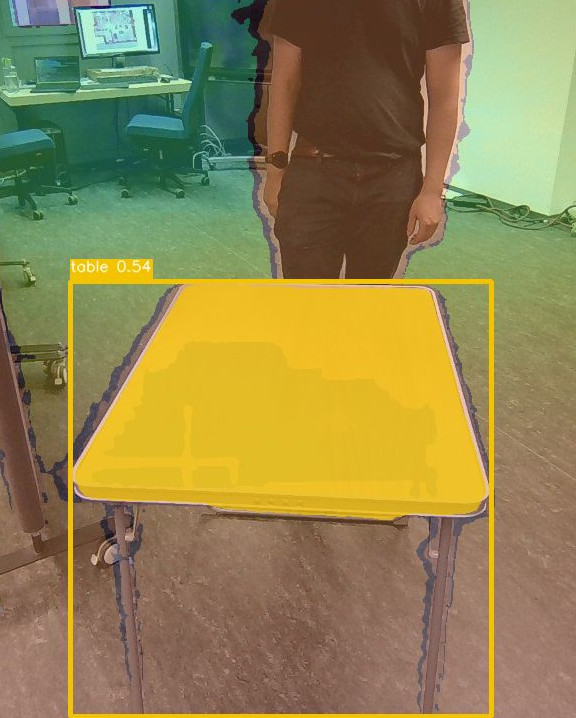}};
  \node[anchor=north west,inner sep=0, xshift=0.2em] (image1) at (image.north east){\includegraphics[height=2.5cm]{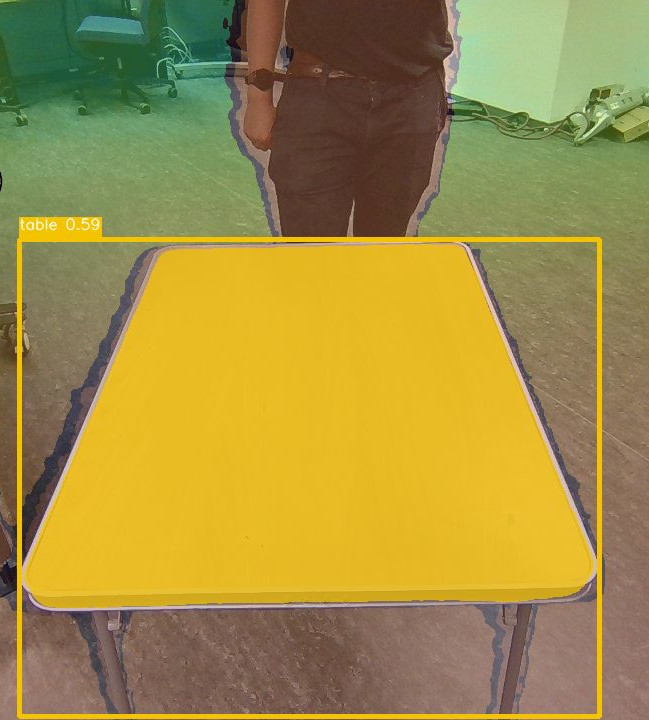}};
  \node[anchor=north west,inner sep=0, xshift=0.2em] (image2) at (image1.north east){\includegraphics[height=2.5cm]{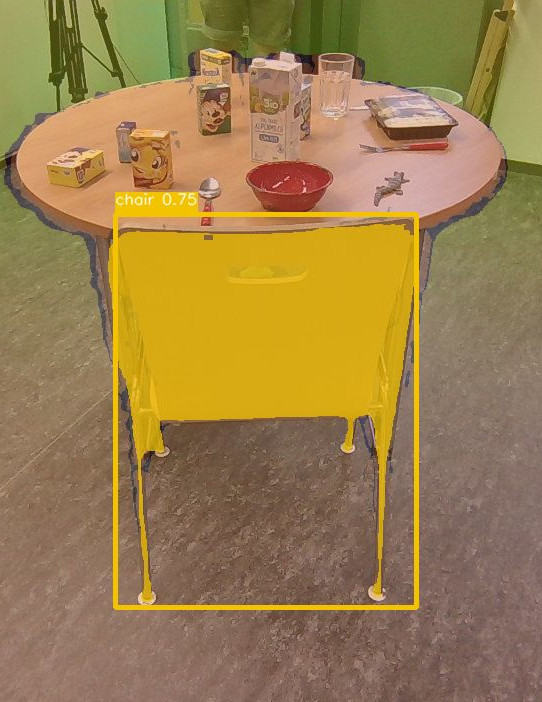}};
  \node[anchor=north west,inner sep=0, xshift=0.2em] (image3) at (image2.north east){\includegraphics[height=2.5cm]{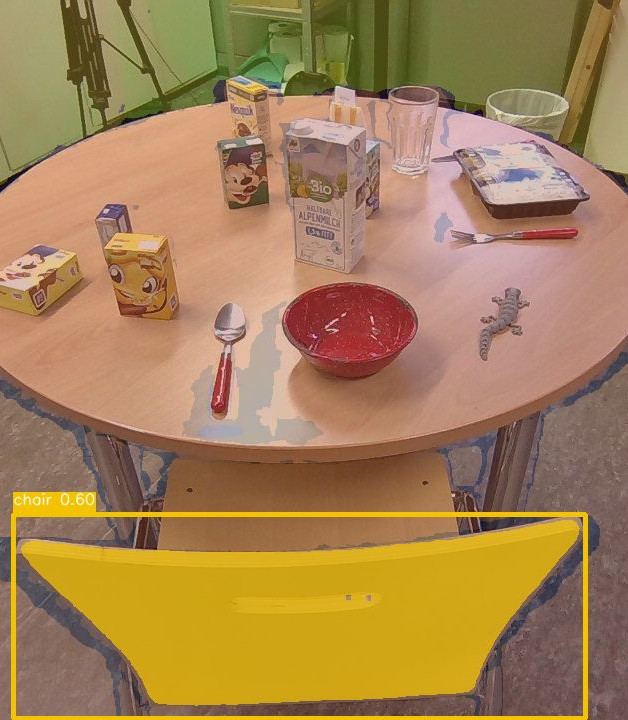}};
  
  \node[inner sep=0.,scale=.9, anchor=north,yshift=-0.3em, rectangle, align=left, font=\scriptsize\sffamily] (n_0) at (image.south) {(a) table before\\\quad\; alignment};
  \node[inner sep=0.,scale=.9, anchor=north,yshift=-0.3em, rectangle, align=left, font=\scriptsize\sffamily] (n_0) at (image1.south) {(b) table after\\\quad\; alignment};
  \node[inner sep=0.,scale=.9, anchor=north,yshift=-0.3em, rectangle, align=left, font=\scriptsize\sffamily] (n_0) at (image2.south) {(c) chair before\\\quad\; alignment};
  \node[inner sep=0.,scale=.9, anchor=north,yshift=-0.3em, rectangle, align=left, font=\scriptsize\sffamily] (n_0) at (image3.south) {(d) chair after\\\quad\; alignment};

  \node[draw, circle, inner sep=0.1em, thick, red,xshift=2.25em, yshift=1.27em] at (image1.south west) {};
  \node[draw, circle, inner sep=0.1em, thick, red,xshift=4.05em, yshift=1.3em] at (image1.south west) {};

  \node[draw, circle, inner sep=0.1em, thick, red,xshift=1.85em, yshift=1.55em] at (image3.south west) {};
  \node[draw, circle, inner sep=0.1em, thick, red,xshift=4.25em, yshift=1.65em] at (image3.south west) {};
  \end{tikzpicture}
  \vspace{-1.8em}
  \caption{Local semantic perception, before and after alignment to table and chair. The red circles denote the computed grasping points.}
  \label{fig:perception_local}
  \vspace{-1.3em}
\end{figure}

\paragraph*{\hspace*{-4.5ex}$\circ$\,Robot}
To plan the robot's arm motions to pick up the targeted furniture objects, the robot uses local, onboard semantic perception from its RGB-D camera.
Upon reaching the goal pose received from the sensor network (cf. Sec.~\ref{sec:nav}), the open vocabulary detector MM-Grounding-Dino~\cite{zhao2024mmgroundingdino} is prompted in a zero-shot manner for tables or chairs in the RGB image.
The resulting object bounding boxes are forwarded to Nano~SAM~\cite{nano_sam}, yielding a segmentation mask per object. 
Since Nano~SAM is a distilled, CNN-based version of Mobile~SAM~\cite{mobile_sam}, a lightweight version of SAM~\cite{kirillov2023sam}, this step adds little computational overhead to the local perception pipeline.
Individual object point clouds are obtained from the depth image using the segmentation masks.
At this point, the closest object is selected for manipulation.
For tables, the tabletop plane is extracted and the corners are sorted to determine the line describing the front side of the table. %
Grasp poses are determined to symmetrically position the grippers parallel around the center of the table front.
For chairs, the backrest is extracted from the point cloud and a line is fitted to its cross-section in the horizontal plane. 
Grasp poses are determined symmetrically around the center of this line and projected into the backrest cross-section. 
Before executing the grasp, the robot base is moved to improve its alignment w.r.t. the estimated object orientation and distance, and the perception is repeated.
\cref{fig:perception_local} shows the local perception of the table and chair before and after alignment of the robot base to the objects.

\subsection{Anticipatory Navigation}
\label{sec:nav}
To realize anticipatory human-aware robot navigation, we integrate semantic feedback about persons and their velocities $\mathcal{P}$, tracked by the smart edge sensor network, into the local dynamic obstacle cost map of the robot navigation stack.
This implementation on a low planning level generalizes well between different robots, and we integrate it into both HSR and TIAGo robots (cf. Sec.~\ref{sec:human_nav}).

The local cost map is a 2D grid map in robot coordinates that incorporates multiple sensor sources (i.e., robot-internal 2D LiDARs and RGB-D camera) that provide information about occupied areas near the robot. It is used to dynamically update the navigation path to avoid obstacles. We add to it a virtual point cloud measurement as an additional sensor source, received as semantic feedback from the sensor network. This virtual point cloud comprises areas predicted to be occupied by the tracked persons.

The virtual point cloud calculation is detailed in Algorithm~\ref{alg:nav}:
We employ the 3D person keypoints $\{\mathbf{kp}\}_{i=1}^{K}$ and their respective linear root joint velocity $\mathbf{v}$ tracked by the sensor network. The joints of persons in the vicinity of the robot are transformed from allocentric to robot-centric coordinates, using the tracked robot pose $\mathbf{p}_\text{r}$. The transformed joints are extrapolated $t_\text{pred}=\SI{2}{\second}$ into the future using the root velocity estimate.
Finally, they are inflated by a safety margin and projected onto the ground plane.
The calculated point cloud $\{\mathbf{kp}_\text{out}\}$ comprising regions that are and will be occupied by persons is sent to the robot for integration into its dynamic obstacle avoidance cost map.

This enables anticipatory human-aware robot navigation, where the robot foresightedly adapts its navigation path e.g., to persons emerging behind occlusions.
Fusing robot internal sensor views with semantic feedback from the ins\-tru\-men\-ted environment allows the robot to ``see around corners''.
\newlength{\textfloatsepsave}
\setlength{\textfloatsepsave}{\textfloatsep}
\setlength{\textfloatsep}{.5em}%
\begin{algorithm}[t]
\caption{Anticipatory Navigation Cost Map}\label{alg:nav}
\hspace*{\algorithmicindent} \textbf{Input:} $t_\text{pred},\,t_\text{step}$ \hfill{\color{gray}\# \textit{prediction time horizon and step}}\\
\hspace*{\algorithmicindent}  $\mathcal{P}=\left\{\left(\left\{\mathbf{kp}\right\}, \mathbf{v}\right)\right\}, \mathbf{p}_{\text{r}}$ \hfill{\color{gray}\# \textit{tracked persons; robot pose}}
\begin{algorithmic}[1]
\State $\{\mathbf{kp}_\text{pred}\} \gets \emptyset$
\State $\mathcal{P}_\text{valid} \gets \text{filter\_outlier}\left(\mathcal{P}\right)$
\Comment{remove far away or incomplete person tracks.}
\ForAll{$P \in \mathcal{P}_\text{valid}$}
\State $\left(\left\{\mathbf{kp}_\text{local}\right\}, \mathbf{v}_\text{local}\right) \gets \text{transform\_frame}\left(P, \mathbf{p}_{\text{r}}\right)$
\State $\{\mathbf{kp}_\text{pred}\} \gets \{\mathbf{kp}_\text{pred}\} \cup  \left\{\mathbf{kp}_\text{local}\right\}$
\For{$\Delta t = 0, t_\text{step}, \dots, t_\text{pred}$}
\State $\{\mathbf{kp}_\text{pred}\} \gets \{\mathbf{kp}_\text{pred}\} \cup  \left\{\mathbf{kp}_\text{local}+\Delta t\cdot\mathbf{v}_\text{local}\right\}$
\EndFor
\EndFor
\State $\left\{\mathbf{kp}_\text{out}\right\} \gets \text{inflate\_and\_proj\_to\_ground}(\{\mathbf{kp}_\text{pred}\})$
\State $\text{publish\_to\_robot}\left(\left\{\mathbf{kp}_\text{out}\right\}\right)$
\Comment{publish point cloud used as additional input to navigation cost map.}
\end{algorithmic}
\end{algorithm}
\begin{algorithm}[t]
\caption{Pickup Pose Anticipation}\label{alg:approach}
\hspace*{\algorithmicindent} \textbf{Input:} $\mathcal{P}=\left\{\mathbf{x}_\text{p}\right\}$ \hfill{\color{gray}\# \textit{tracked persons (root kps only)}}\\
\hspace*{\algorithmicindent} $\mathcal{T}=\{\left[\mathbf{x}_\text{t},\theta_\text{t}\right]{}^\top\}, \mathcal{A}_\text{pick}$\hfill{\color{gray}\# \textit{tracked tables; pickup area}}\\
\hspace*{\algorithmicindent} $\mathbf{p}_\text{r}, \mathbf{v}_\text{r}, \delta_\text{grasp}$ \hfill{\color{gray}\# \textit{robot pose \& velocity; grasp distance}}\\
\hspace*{\algorithmicindent} $\tau_\text{p}, \tau_\text{goal}, \tau_\text{vel}$ \hfill{\color{gray}\# \textit{pickup and goal pos \& vel. thresholds}}
\begin{algorithmic}[1]
\While{True}
\State $\mathcal{T}_\text{pick} \gets \text{select\_tables}\left(\mathcal{T}, \mathcal{A}_\text{pick}\right)$
\Comment{tables in pick area.}
\State $P \gets \text{select\_person}\left(\mathcal{P}, \mathcal{A}_\text{pick}, \tau_\text{p}\right)$
\Comment{person closest to pickup area center and distance within $\tau_\text{p}$.}
\If{$P$ not None}
\State $T \gets \text{closest\_table}\left(\mathcal{T}_\text{pick}, P\right)$
\LComment{Calculate goal pose to grasp table from opposite side than person.}
\State $\mathbf{n}_\text{t} \gets \left[\cos\left(\theta_\text{t}\right), \sin\left(\theta_\text{t}\right)\right]{}^\top $
\Comment{table long side.}
\State $\mathbf{n}_\text{p} \gets \mathbf{x}_\text{t} - \mathbf{x}_\text{p}$
\Comment{person approach vector.}
\State $\mathbf{x}_\text{goal} \gets \mathbf{x}_\text{t} + \sign\left(\mathbf{n}_\text{p}^\top \mathbf{n}_\text{t}\right) \delta_\text{grasp} \mathbf{n}_\text{t}$
\State $\mathbf{p}_\text{goal} \gets \left[\mathbf{x}_\text{goal}, \text{adjust\_ori}\left(\theta_\text{t}, \sign\left(\mathbf{n}_\text{p}^\top \mathbf{n}_\text{t}\right)\right)\right]$
\State $\text{publish\_to\_robot}\left(\mathbf{p}_\text{goal}\right)$
\Comment{send via network.}
\If{$\text{pos\_diff}\left(\mathbf{p}_\text{goal}, \mathbf{p}_\text{r}\right) < \tau_\text{goal}$ and $\lvert\mathbf{v}_\text{r}\rvert < \tau_\text{vel}$}
\State \Return
\EndIf
\EndIf
\EndWhile
\end{algorithmic}
\end{algorithm}

For collaborative furniture handling, we anticipate the human's intention for where to pick up and place the furniture based on a pre-defined pickup area~$\mathcal{A}_\text{pick}$ and target layout~$\mathcal{L}$.
The pickup pose anticipation is detailed in Algorithm~\ref{alg:approach}.
When the sensor network detects a person $P$ moving towards $\mathcal{A}_\text{pick}$, the robot receives a goal pose $\mathbf{p}_\text{goal} \in \mathbb{SE}(2)$ to navigate to the table $T$ closest to the person. This goal pose is dynamically adjusted depending on the table side the person moves to, or if they move on to another table.
The goal position $\mathbf{x}_\text{goal}$ is offset from the table center position $\mathbf{x}_\text{t}$ by the grasp distance $\delta_\text{grasp}$ along the table's x-axis $\mathbf{n}_\text{t}$ towards the opposite side the person is approaching from. The orientation is adjusted so that the robot faces the table. This ensures the robot aligns correctly for collaborative grasping. When the goal pose is reached, the robot base alignment is refined based on the robot's onboard perception and the object is grasped as detailed in Sec.~\ref{sec:perception} and~\ref{sec:manip}.

\subsection{Collaborative Manipulation}
\label{sec:manip}

Given a viable alignment of the robot base, a bimanual grasping motion is performed by bringing the robot arms forward from their home configuration and moving the end-effectors to pre-grasp orientations for tables or chairs using predefined joint space commands. %
Next, a two-step Cartesian trajectory is computed and executed based on the grasp poses calculated by the local semantic perception.
This trajectory moves the end-effectors to their grasp poses while ensuring that the final approach is exclusively along the open direction of the grippers to prevent collisions with the object surface.

\begin{algorithm}[t]
\caption{Collaborative Carrying and Goal Anticipation}\label{alg:followhand}
\hspace*{\algorithmicindent} \textbf{Input:} $\mathbf{x}_\text{ee,init}, \mathbf{x}_\text{ee}$\hfill{\color{gray}\# \textit{init. \& curr. end-effector position}}\\
\hspace*{\algorithmicindent} $\mathcal{L}=\{\left[\mathbf{x}_\text{t},\theta_\text{t}\right]{}^\top\}, \mathbf{p}_\text{r}=\left[\mathbf{x}_\text{r},\theta_\text{r}\right]{}^\top$ \hfill{\color{gray}\# \textit{layout; robot pose}}\\
\hspace*{\algorithmicindent} $\tau_\text{ee}, \tau_\text{goal}, \tau_\text{direct}$ \hfill{\color{gray}\# \textit{thresholds}}\\
\hspace*{\algorithmicindent} $k_\text{e,lin}, k_\text{e,rot}, k_\text{a,lin}, k_\text{a,rot}, k_\text{direct}$ \hfill{\color{gray}\# \textit{scaling factors}}
\begin{algorithmic}[1]
\While{True}
\State $\mathbf{p}_\text{goal} \gets \text{update\_goal}\left(\mathcal{L}, \mathbf{p}_\text{r}\right)$
\Comment{robot pose to place table at layout pose closest to current position.}
\State $\text{look\_at}\left(\mathbf{p}_\text{goal}\right)$
\Comment{robot head looks at goal.}
\State $\mathbf{\Delta x}_\text{ee} \gets \mathbf{x}_\text{ee} - \mathbf{x}_\text{ee,init}$
\Comment{end-effector displacement.}
\State $\mathbf{\Delta x}_\text{goal} \gets \mathbf{x}_{\text{goal}} - \mathbf{x}_{\text{r}}$
\Comment{vector to goal.}
\State $\Delta \theta_\text{goal} \gets \text{angle\_diff}\left(\theta_{\text{goal}} - \theta_{\text{r}}\right)$ 
\State $\mathbf{v}_\text{ee} \gets \mathbf{0}, \mathbf{v}_\text{a} \gets \mathbf{0}, \mathbf{v}_\text{r} \gets \mathbf{0}$
\Comment{$\mathbf{v} = \left[v_x, v_y, \omega_\theta\right]{}^\top.$}
\If{$\lvert\lvert\mathbf{\Delta x}_\text{goal}\rvert\rvert < \tau_\text{goal}$}
\Comment{goal reached.}
\State $\text{publish}\left(\mathbf{v}_\text{r}\right)$
\State \Return
\EndIf
\If{$\lvert\Delta x_{\text{ee}}\rvert > \tau_\text{ee}$}
\Comment{forward movement.}
\State $v_{\text{ee},x} \gets k_\text{e,lin} \cdot \Delta x_{\text{ee}}$
\EndIf
\If{$\lvert\Delta y_{\text{ee}}\rvert > \tau_\text{ee}$}
\Comment{lateral movement or rotation.}
\If{$\lvert\Delta x_{\text{ee}}\rvert > \tau_\text{ee}$}
\State $\omega_{\text{ee},\theta} \gets k_\text{e,rot} \cdot \Delta y_{\text{ee}}$
\Comment{when moving forward, use lateral $\Delta y_{\text{ee}}$ for turning.}
\Else
\Comment{else, use lateral $\Delta y_{\text{ee}}$ for sideways motion.}
\State $v_{\text{ee},y} \gets k_\text{e,lin} \cdot \Delta y_{\text{ee}}$
\EndIf
\EndIf
\State $\mathbf{v}_{\text{a},xy} \gets k_\text{a,lin} \cdot \mathbf{\Delta x}_\text{goal}$
\Comment {linear velocity to goal.}
\State $\omega_{\text{a},\theta} \gets k_\text{a,rot} \cdot \Delta \theta_\text{goal}$
\Comment {angular velocity to goal.}
\If{$\lvert\lvert\mathbf{\Delta x}_\text{goal}\rvert\rvert > \tau_\text{direct}$ or $\lvert\Delta x_{\text{ee}}\rvert > \tau_\text{ee}$}
\State $\mathbf{v}_\text{r} \gets \mathbf{v}_\text{ee} + \mathbf{v}_\text{a}$
\Comment{far from goal or moving: combine end-effector and goal velocities}
\Else
\Comment{close to goal and stopped: direct approach.}
\State $\mathbf{v}_\text{r} \gets k_\text{direct} \cdot \mathbf{v}_\text{a}$
\Comment{scale up vel. due to small diffs.}
\EndIf
\State $\text{publish}\left(\mathbf{v}_\text{r}\right)$
\EndWhile
\end{algorithmic}
\end{algorithm}

Once the table is grasped, the robot anticipates the human's intention to lift it and activates a compliant control mode by switching its arms to a gravity-compensation mode, configured to hold the weight of the table.
The person on the other side of the table can move the robot's arms through it to enable collaborative carrying of the furniture.

Velocity commands $\mathbf{v}_\text{ee}= [v_x, v_y, \omega_\theta]^\top$ for both linear and angular robot base movements are calculated proportionally to the end-effector displacement $\mathbf{\Delta x}_\text{ee} = [\Delta x_{\text{ee}}, \Delta y_{\text{ee}}]^\top$ as detailed in Algorithm~\ref{alg:followhand}.
A forward displacement $\Delta x_{\text{ee}}$ larger than $\tau_\text{ee}$ will induce forward motion, and a sideways displacement $\Delta y_{\text{ee}}$ larger than $\tau_\text{ee}$ a turn when moving forward, or a sideways movement when there is no forward movement (Alg.~\ref{alg:followhand}, ll.~11-17). This enables the person to control all degrees of freedom of the omnidirectional robot base for guiding the robot toward the target pose.

The robot further anticipates the target pose to place the table from the furniture layout $\mathcal{L}$.
The robot head turns to look at the selected goal to signal the anticipation to the operator.
The velocity $\mathbf{v}_\text{ee}$ is overlaid with a velocity vector $\mathbf{v}_\text{a}$ pointing towards the anticipated goal (Alg.~\ref{alg:followhand}, ll.~18-21). 
Thus, the robot automatically draws towards the placement location but can still be influenced by the person, e.g., to avoid obstacles or towards a different target pose from $\mathcal{L}$.
Fig.~\ref{fig:robot_human} illustrates the collaborative carrying.

Once the target comes closer than a threshold $\tau_\text{direct}=\SI{1}{\meter}$ and the operator stops the robot by moving the end-effectors back to the initial position $\mathbf{x}_\text{ee,init}$, the robot moves the last part fully automatically (Alg.~\ref{alg:followhand} ll.~22-23), to achieve a precise positioning of the furniture. When the target pose is reached within tolerance $\tau_\text{goal}$, the robot stops the compliant mode and executes the placement motion, which is the reverse of the grasp motion.
After placement, the now occupied target pose is removed from the layout plan $\mathcal{L}$, and the robot moves on to pick up the next object.

\def\robotpos{(5.6,0)}
\def\robotsize{1.5}
\def\tablecenter{(3.83,.6)}
\def\tabledirection{(-0.5,0.0)}
\def\robotdirection{(-0.5,0)}

\def\goaltablepos{(2,3)}
\def\goaltableposcenter{(2.5,4)}
\def\goaltabledirection{(0,0.5)}

\def\humanpos{(2.35,1.)}
\def\humandirection{(-1.0,0)}

\def\endeffectorcenter{(-1.,0.2)}

\def\robottargetpos{(2.5,2.5)}

\def\centerarc[#1](#2)(#3:#4:#5){ \draw[#1] ($(#2)+({#5*cos(#3)},{#5*sin(#3)})$) arc (#3:#4:#5); }

\tikzset{
  robot/.pic={
    \draw[thick, fill=red!20] (0,0) circle (0.5);
    \draw[thick, fill=black] (0,0) circle (0.05);

    \draw[thick, color=black] (0.3,.1) rectangle ++ (0.15,0.75);
    \draw[thick, color=black] (0.25,0.75) ++ (0.0,0.1) rectangle ++ (0.25,0.25); %

    \draw[thick, color=black] (-0.2,.1) ++ (-0,0.) rectangle ++ (0.15,0.75);
    \draw[thick, color=black] (-0.2,0.1) ++ (-0.05,0.0) ++ (0,0.75) rectangle ++ (0.25,0.25); %
    \draw[thick, fill=black] (-0.25,0.85) ++ (0.125,0.125) circle (0.05) node[anchor=center] (ee1) {}; %

    \draw[thick, color=blue] (0.05,.1) rectangle ++ (0.15,0.45);
    \draw[thick, color=blue] (0.1,0.45) ++ (-0.1,0.1) rectangle ++ (0.25,0.25);

    \draw[thick, color=blue] (-0.2,.1) ++ (-0.25,0.) rectangle ++ (0.15,0.45);
    \draw[thick, color=blue] (-0.35,0.45) ++ (-0.15,0.1) rectangle ++ (0.25,0.25);
    \draw[thick, fill=blue, color=blue] (-0.5,0.55) ++ (0.125,0.125) circle (0.05) node[anchor=center] (ee2) {};%

    \draw[<-, thick, color=red, dotted] (ee1.center) -- (ee2.center) node[midway] (midee) {};%

    \draw[ -, dashed] (ee1.center) -- ++(-0.5,0.0) node[anchor=north] {$\mathbf{x}_\text{ee}$};
    \draw[ -, dashed, color=blue] (ee2.center) -- ++(-0.4,-0.1) node[anchor=west, yshift=-1., xshift=-1.] {$\mathbf{x}_\text{ee,init}$};
    \draw[ -, dashed, color=red] (midee) -- ++(-0.5,0.8) node[anchor=north] {$\mathbf{\Delta x}_\text{ee}$};

    \node[anchor=north, xshift=2.] at (0,0) {$\mathbf{p}_{\text{r}}$};
    \draw[thick, ->] (0,0) -- (0,0.5);
  }
}

\begin{figure}[t]
  \centering

  \begin{tikzpicture}

    \draw[thick, gray] (0.,5.15) rectangle ++ (8,-6.);

    \draw[thick, dashed, fill=green!20] \goaltablepos rectangle ++ (1,2);
    \draw[thick, dashed, ->] \goaltableposcenter -- ++ \goaltabledirection;
    \draw[thick, fill=black] \goaltableposcenter circle (0.05);

    \draw[thick, dashed, pattern color=green!20, pattern=crosshatch] \goaltablepos ++ (-1.2,0) rectangle ++ (1,2);
    \draw[thick, fill=black] \goaltableposcenter ++ (-1.2,0) circle (0.05);
    \draw[thick, dashed, ->] \goaltableposcenter ++ (-1.2,0) -- ++ \goaltabledirection;

    \draw[thick, fill=red!20, dashed] \robottargetpos circle (0.5);
    \draw[thick, fill=black] \robottargetpos circle (0.05) node[below] {$\mathbf{p}_{\text{goal}}$};
    \draw[thick, dashed, ->] \robottargetpos -- ++ (0,0.5);
    \draw[thick, ->] \robottargetpos -- ++ (-0.483,0.1294); %
    \centerarc[->,red,thick, dotted](2.5,2.5)(165:90:0.3); %
    \draw[ -, dashed, red] \robottargetpos ++ (-0.13,0.13) -- ++(1.,0.0) node[anchor=west] {$\Delta \theta_\text{goal}$};

    \draw[thick, fill=yellow!20] \humanpos circle (0.5);
    \draw[thick, fill=black] \humanpos circle (0.05) node[below] {$\mathbf{x}_{\text{p}}$};
    \draw[thick, fill=orange!20, rotate around={-15:\tablecenter}] \tablecenter ++ (-1,-0.5) rectangle ++ (2,1);
    \draw[thick,->, rotate around={-15:\tablecenter}] \tablecenter -- ++ \tabledirection;
    \draw[thick, fill=black] \tablecenter circle (0.05);
    \pic[rotate=75] at \robotpos {robot};
    \draw[dotted, thick, color=red,->] \robotpos -- \robottargetpos node[midway, above] {$\mathbf{\Delta x}_{\text{goal}}$};
  \end{tikzpicture}
  \vspace{-1.0em}
  \caption{Collaborative carrying with goal anticipation including robot pose $\mathbf{p}_{\text{r}}$, human position $\mathbf{x}_{\text{p}}$, the target table layout (green), anticipated goal pose $\mathbf{p}_{\text{goal}}$, and the end-effector initial and current positions $\mathbf{x}_{\text{ee, init}}$ resp. $\mathbf{x}_{\text{ee}}$.
  The robot velocity is determined w.r.t. the position and orientation difference to the goal $\mathbf{\Delta x}_{\text{goal}}$ resp. $\Delta\theta_{\text{goal}}$, and the end-effector displacement $\mathbf{\Delta x}_{\text{ee}}$.}
  \label{fig:robot_human}
  \vspace{-1.em}
\end{figure}

The chairs, in contrast to the tables, are handled by the robot alone, without compliant movement control, directly pulling and pushing them from pickup to placement location.

\section{Experimental Evaluation}
\subsection{Setup}

\begin{figure}[t]
  \centering
\begin{tikzpicture}[boxstyle/.style={font=\small\sffamily,black,fill=blue!20!white,fill opacity=0.8,text opacity=1,text=black,draw,very thick,align=center,rectangle callout}]
          \definecolor{red}{rgb}{0.7,0.0,0.0}
        \definecolor{blue}{rgb}{0.2,0.2,0.7}
    \node[anchor=north west,inner sep=0] (image) at (0,0) {\includegraphics[width=0.3\columnwidth]{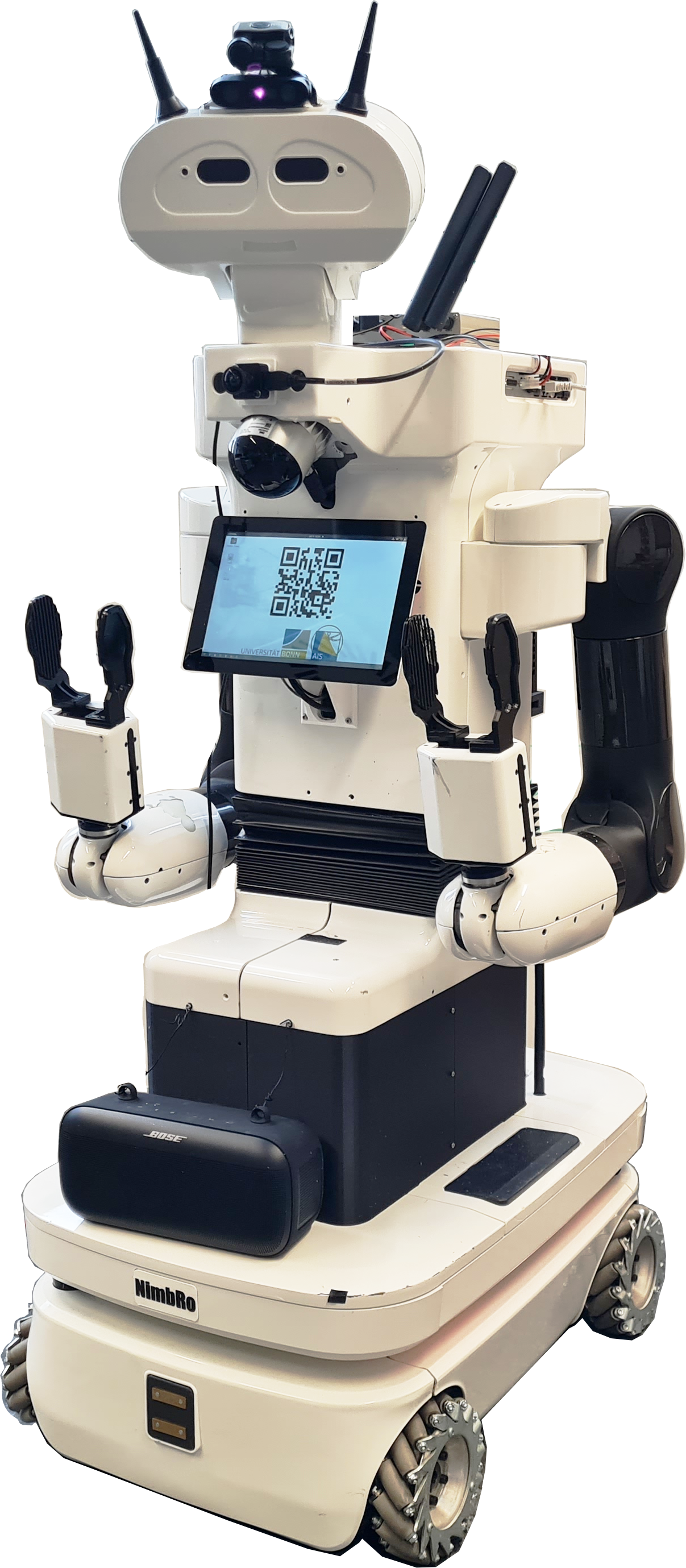}};
    \node[label,scale=.9, anchor=south west, xshift=-0.55cm, rectangle, align=center, font=\footnotesize\sffamily] (n_0) at (image.south west) {(a)};

    \begin{scope}[shift=(image.south west),x={(image.south east)},y={(image.north west)}]
      \node[boxstyle, line width=0.1mm, scale=0.6,text width=3.2cm,callout relative pointer={(0.6, 0)}] at (-0.45,0.95) {Orbbec Gemini~335 RGB-D camera};
      \node[boxstyle, line width=0.1mm, scale=0.6,text width=2.5cm,callout relative pointer={(0.35, 0)}] at (-0.55,0.55) {Parallel grippers};
      \node[boxstyle, line width=0.1mm, scale=0.6,text width=2.5cm,callout relative pointer={(0.35, 0)}] at (-0.55,0.45) {7-DoF arms};
      \node[boxstyle, line width=0.1mm, scale=0.6,text width=3.cm,callout relative pointer={(0.35, -0.02)}] at (-0.55,0.15) {Omnidirectional base};
      \node[boxstyle, line width=0.1mm, scale=0.6,text width=2.5cm,callout relative pointer={(0.75, -0.02)}] at (-0.55,0.65) {Touchscreen};
      \node[boxstyle, line width=0.1mm, scale=0.6,text width=3.cm,callout relative pointer={(0.65, -0.1)}] at (-0.45,0.8) {Zotac ZBOX QTG7A4500 (back)};
    \end{scope}

    \node[anchor=north west,inner sep=0] (sensoredge) at (4.,0) {\includegraphics[width=0.25\columnwidth, trim=0 0 200px 0, clip]{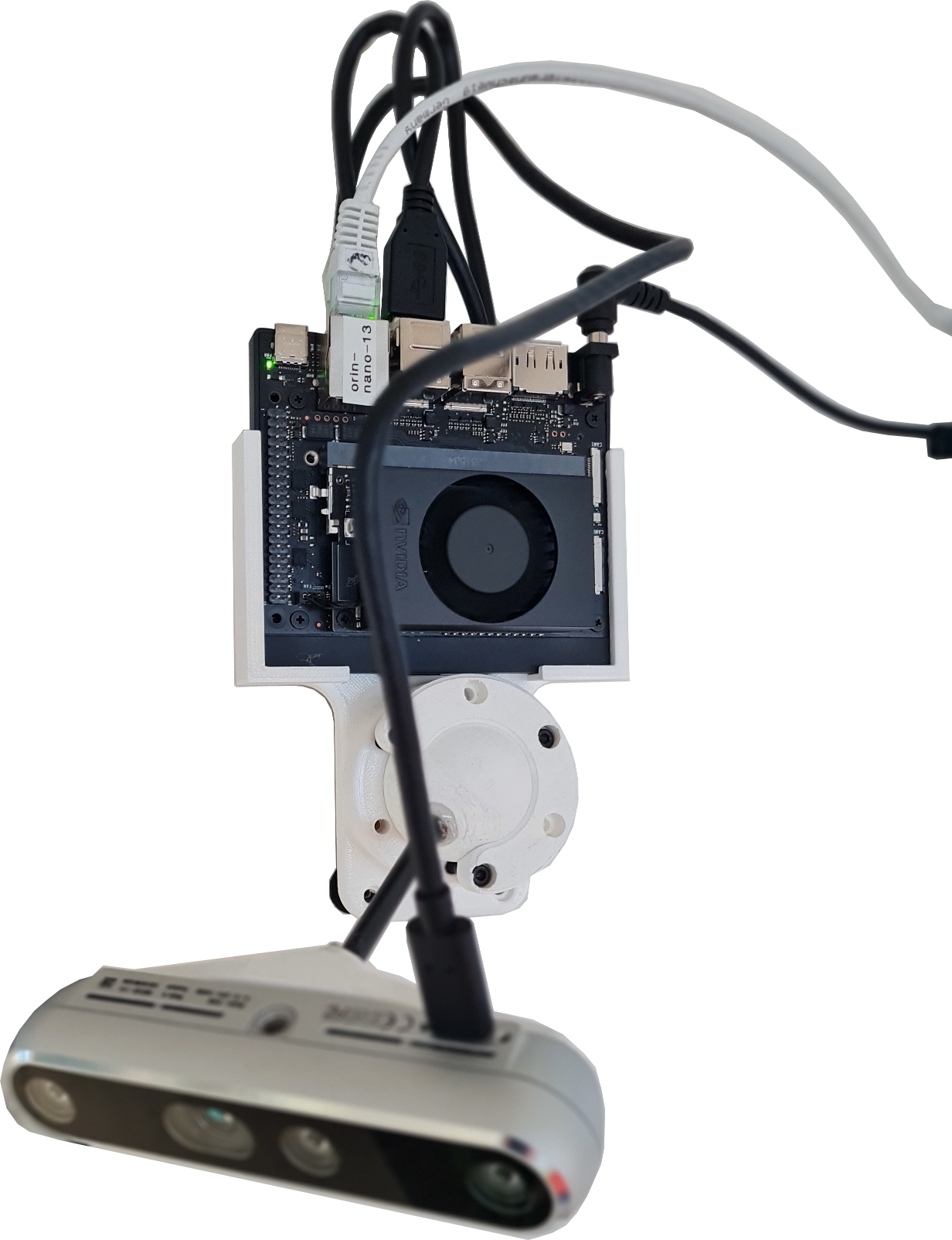}};
    
    \node[label,scale=.9, anchor=south west,xshift=-0.5cm, rectangle, align=center, font=\footnotesize\sffamily] (n_1) at (n_0.south -| sensoredge.south west) {(b)};
    \begin{scope}[shift=(sensoredge.south west),x={(sensoredge.south east)},y={(sensoredge.north west)}]
      \node[boxstyle, line width=0.1mm, scale=0.6,text width=2.2cm,callout relative pointer={(0.35, -0.07)}] at (-.15,0.75) {Nvidia Jetson Orin Nano};
      \node[boxstyle, line width=0.1mm, scale=0.6,text width=2.2cm,callout relative pointer={(0.25, -0.2)}] at (-.15,0.45) {RealSense D455 \mbox{RGB-D} camera};
    \end{scope}
  \end{tikzpicture}
   \vspace{-1.5em}
  \caption{Robot and sensor setup: (a) PAL Robotics TIAGo++ robot; (b) sample \smartedge sensor.}
  \label{fig:setup}
  \vspace{-1.5em}
\end{figure}

For our evaluation experiments, we use a PAL Robotics TIAGo++ \cite{pages2016tiago} omnidirectional dual-arm mobile manipulator, shown in \cref{fig:setup}\,(a), equipped with an Orbbec Gemini~335 RGB-D camera and a Zotac ZBOX QTG7A4500 computer mounted on the back of the robot.

The robot is informed by a \smartedge sensor network consisting of 25 sensor nodes, shown in \cref{fig:setup}\,(b), with an Intel RealSense D455 RGB-D camera and an Nvidia Jetson Orin Nano compute board with an embedded GPU for onboard semantic perception using lightweight CNNs~\citep{bultmann_ias2022,hau_objectlevel_2022}. The smart edge sensors are mounted at a height of $\sim$\SI{2.5}{\meter}, distributed over a lab space of $\sim$\SI{240}{\square\meter} size.

\subsection{Evaluation}
Many approaches related to furniture manipulation focus on qualitative demonstrations. 
In this work, we conduct repetitive experiments following a task protocol for both anticipatory navigation and collaborative furniture arrangement.
\subsubsection{Human-Aware Anticipatory Navigation}
\label{sec:human_nav}

In our anticipatory human-aware robot navigation experiment, we use the real-time 3D human pose tracking by the external smart edge sensors to send semantic feedback informing the robot about people in its vicinity but out of sight of its internal sensors, e.g., due to occlusions or limited \ac{FoV} (cf. Sec.~\ref{sec:nav}).
This scenario is common in many household or office environments, e.g., at corridor intersections, as well as for warehouses with narrow aisles between high shelves, where people suddenly emerging from behind occlusions can be at risk of collision with autonomously operating robots.

\setlength{\textfloatsep}{\textfloatsepsave}

\begin{figure*}[t]
  \centering
  \begin{tikzpicture}
  \node[anchor=north west,inner sep=0] (image) at (0,0){\includegraphics[height=3.05cm]{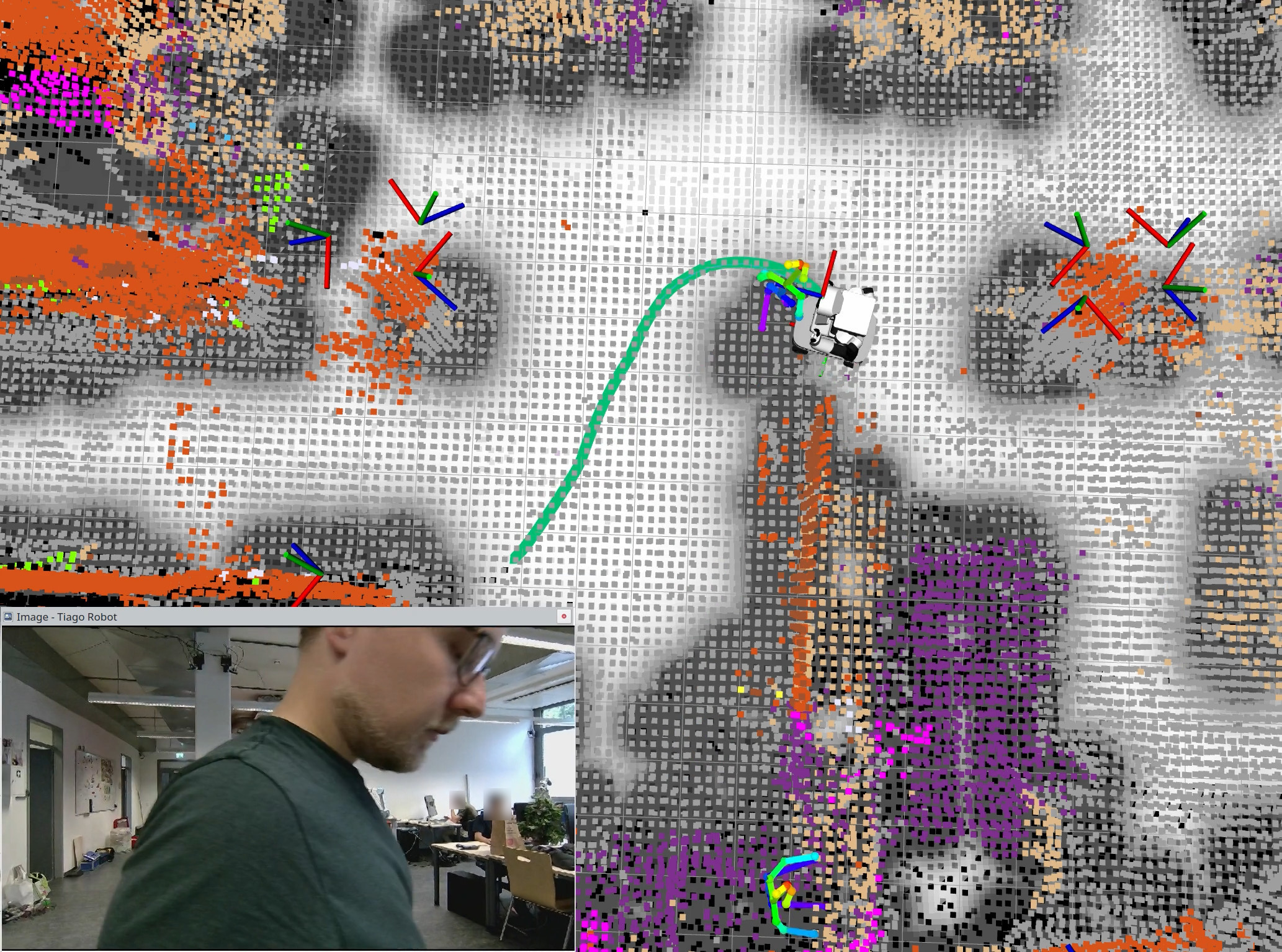}};
  \node[anchor=north west,inner sep=0, xshift=0.2em] (image2) at (image.north east){\includegraphics[height=3.05cm]{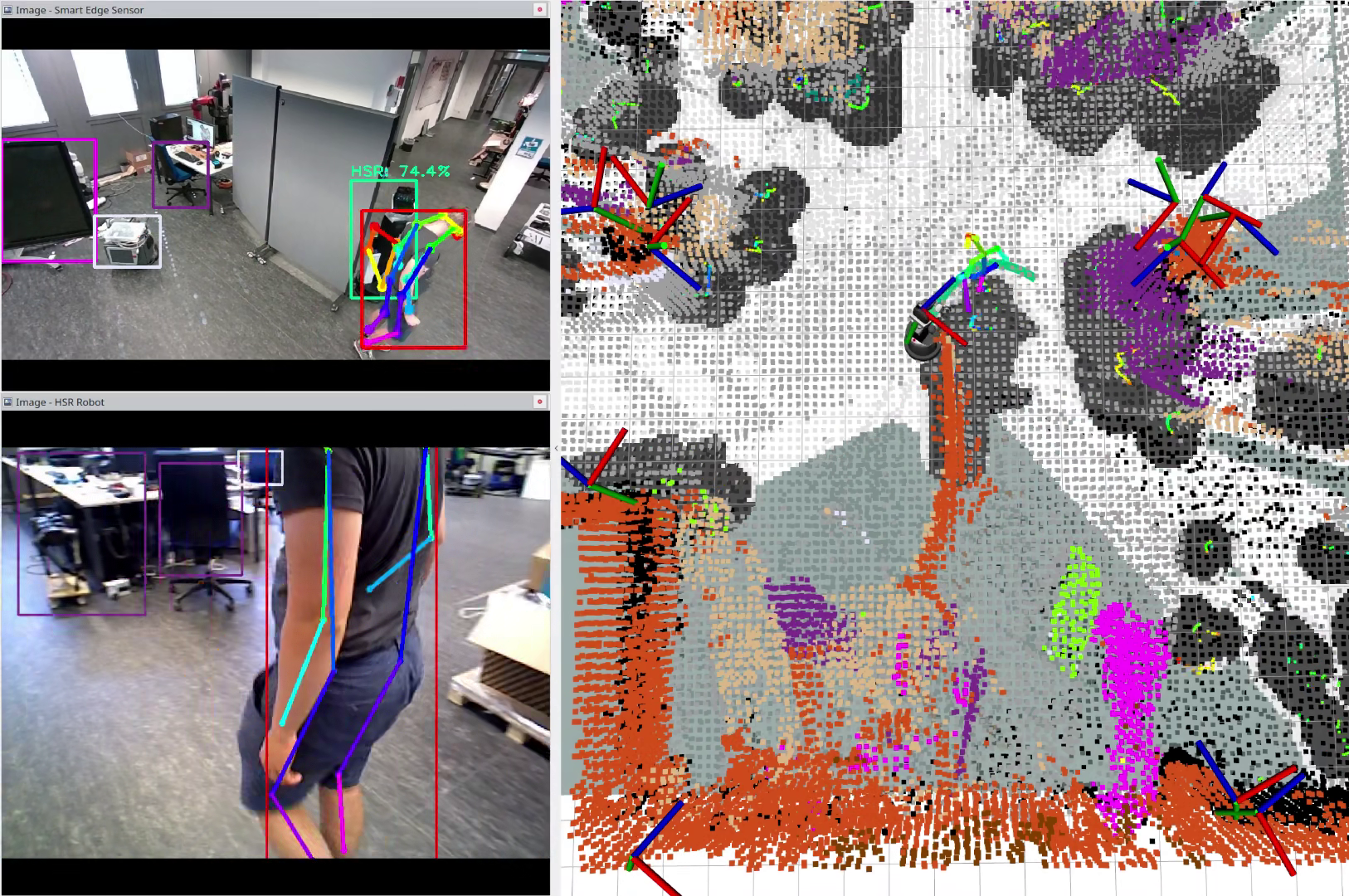}};
  
  \node[anchor=north west,inner sep=0, xshift=0.5em] (image1) at (image2.north east){\includegraphics[height=3.05cm]{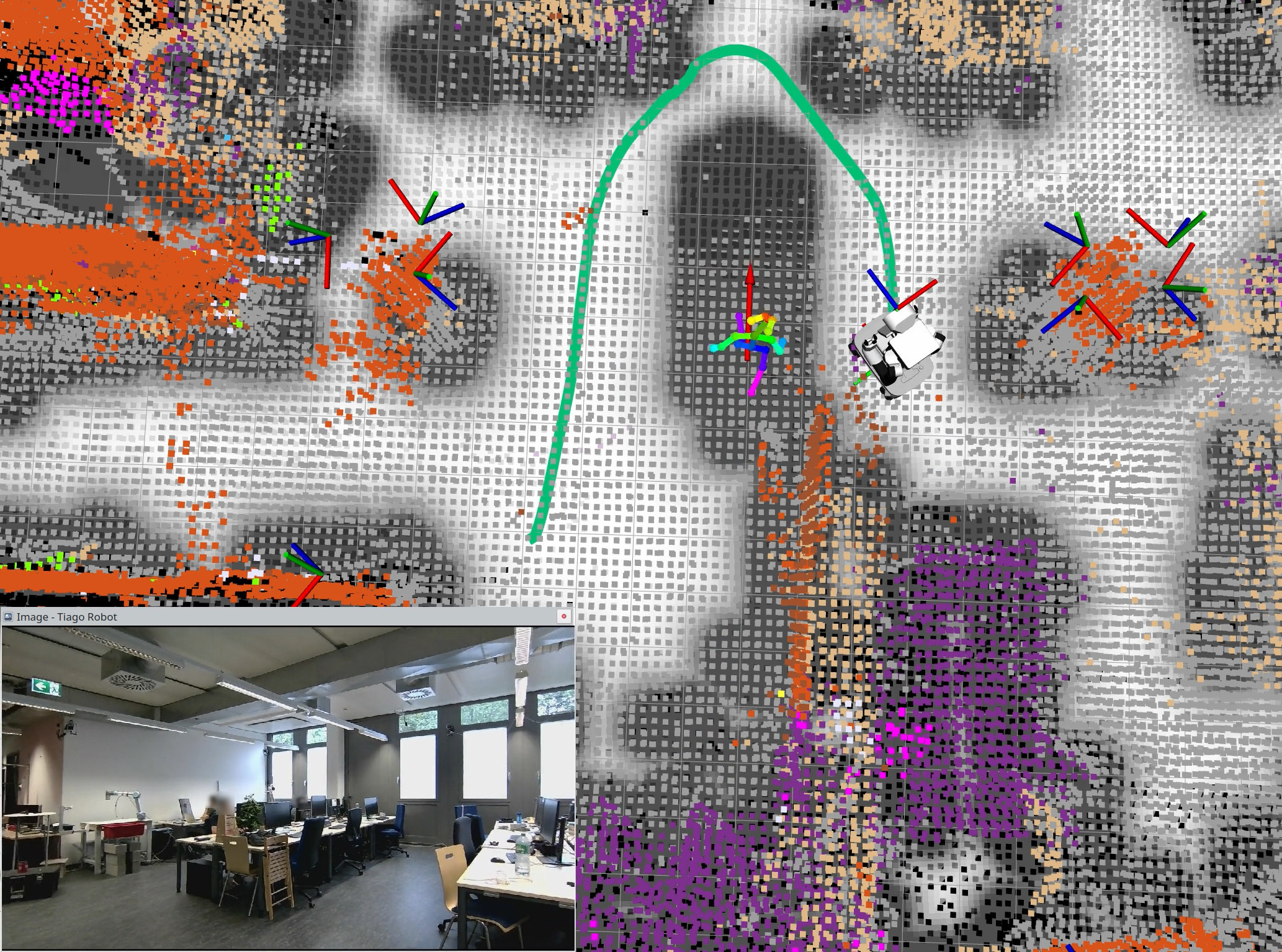}};
  
  \node[anchor=north west,inner sep=0, xshift=0.2em] (image3) at (image1.north east){\includegraphics[height=3.05cm]{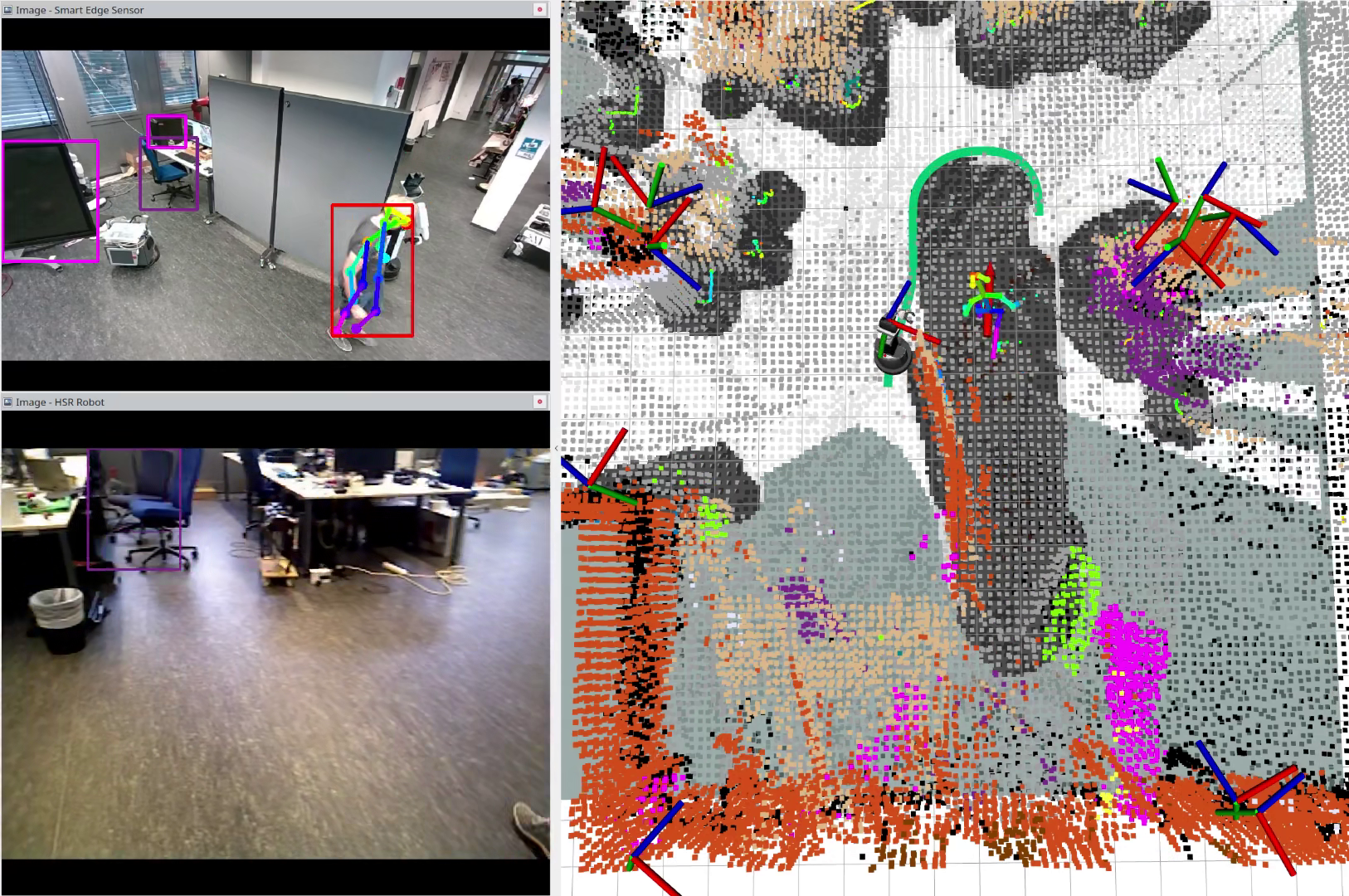}};
  
  \node[inner sep=0.,scale=.9, anchor=north,yshift=-0.3em, rectangle, align=left, font=\scriptsize\sffamily] (n_0) at (image2.south west) {(a) without anticipation: risk of collision with person};
  \node[inner sep=0.,scale=.9, anchor=north,yshift=-0.3em, rectangle, align=left, font=\scriptsize\sffamily] (n_0) at (image3.south west) {(b) with anticipation: foresighted person avoidance};
  
  \begin{scope}[shift=(image.north west),x={(image.north east)},y={(image.south west)}]
			\node[label, scale=.5, anchor=south west, rectangle, rounded corners=2, inner sep=0.06cm, fill={white!90!black},opacity=.9,text opacity=1, align=center, font=\scriptsize\sffamily] (l_cam2) at (0.27, 0.17) {Cam\,1};
    		\draw[{white!90!black},opacity=.9, very thick] (l_cam2.300) ++(0., -0.002) -- ++(0., 0.05);
    				    
		    \node[label, scale=.5, anchor=south west, rectangle, rounded corners=2, inner sep=0.06cm, fill={white!90!black},opacity=.9,text opacity=1, align=center, font=\scriptsize\sffamily] (l_cam1) at (0.265, 0.385) {Cam\,2};
			\draw[{white!90!black},opacity=.9, very thick] (l_cam1.60) ++(0., 0.002) -- ++(0., -0.04);
    		
    		\node[label, scale=.5, anchor=south west, rectangle, rounded corners=2, inner sep=0.06cm, fill={white!90!black},opacity=.9,text opacity=1, align=center, font=\scriptsize\sffamily] (l_cam4) at (0.82, 0.42) {Cam\,3};
    		\draw[{white!90!black},opacity=.9, very thick] (l_cam4.130) ++(0., 0.002) -- ++(-0., -0.04);
    		
    		\node[label, scale=.5, anchor=south west, rectangle, rounded corners=2, inner sep=0.06cm, fill={white!90!black},opacity=.9,text opacity=1, align=center, font=\scriptsize\sffamily] (l_cam3) at (0.82, 0.195) {Cam\,4};
    		\draw[{white!90!black},opacity=.9, very thick] (l_cam3.225) ++(0., -0.002) -- ++(0., 0.05);
    		
    		 \node[label, scale=.5, anchor=south west, rectangle, rounded corners=2, inner sep=0.06cm, fill={white!90!black},opacity=.9,text opacity=1, align=center, font=\scriptsize\sffamily] (l_camR) at (0.68, 0.34) {Robot\\Cam};
    \draw[{white!90!black},opacity=.9, very thick] (l_camR.190) ++(0.002, 0.) -- ++(-0.03, 0.);
    		
			\node[label,scale=.5, anchor=north west, rectangle, fill=white, align=center, font=\scriptsize\sffamily] (n_1) at (0,0.64) {TIAGo Robot Cam};
			\node[label,scale=.5, anchor=south west, rectangle, fill=white, align=center, font=\scriptsize\sffamily] (n_2) at (0.45, 1.) {3D Scene View};
		
    	\end{scope}
    	
   		\begin{scope}[shift=(image1.north west),x={(image1.north east)},y={(image1.south west)}]
			\node[label, scale=.5, anchor=south west, rectangle, rounded corners=2, inner sep=0.06cm, fill={white!90!black},opacity=.9,text opacity=1, align=center, font=\scriptsize\sffamily] (l_cam2) at (0.27, 0.17) {Cam\,1};
    		\draw[{white!90!black},opacity=.9, very thick] (l_cam2.300) ++(0., -0.002) -- ++(0., 0.05);
    				    
		    \node[label, scale=.5, anchor=south west, rectangle, rounded corners=2, inner sep=0.06cm, fill={white!90!black},opacity=.9,text opacity=1, align=center, font=\scriptsize\sffamily] (l_cam1) at (0.265, 0.385) {Cam\,2};
			\draw[{white!90!black},opacity=.9, very thick] (l_cam1.60) ++(0., 0.002) -- ++(0., -0.04);
    		
    		\node[label, scale=.5, anchor=south west, rectangle, rounded corners=2, inner sep=0.06cm, fill={white!90!black},opacity=.9,text opacity=1, align=center, font=\scriptsize\sffamily] (l_cam4) at (0.82, 0.42) {Cam\,3};
    		\draw[{white!90!black},opacity=.9, very thick] (l_cam4.130) ++(0., 0.002) -- ++(-0., -0.04);
    		
    		\node[label, scale=.5, anchor=south west, rectangle, rounded corners=2, inner sep=0.06cm, fill={white!90!black},opacity=.9,text opacity=1, align=center, font=\scriptsize\sffamily] (l_cam3) at (0.82, 0.195) {Cam\,4};
    		\draw[{white!90!black},opacity=.9, very thick] (l_cam3.225) ++(0., -0.002) -- ++(0., 0.05);
    		
    		 \node[label, scale=.5, anchor=south west, rectangle, rounded corners=2, inner sep=0.06cm, fill={white!90!black},opacity=.9,text opacity=1, align=center, font=\scriptsize\sffamily] (l_camR) at (0.73, 0.34) {Robot\\Cam};
    \draw[{white!90!black},opacity=.9, very thick] (l_camR.195) ++(0.002, 0.) -- ++(-0.03, 0.);
    		
			\node[label,scale=.5, anchor=north west, rectangle, fill=white, align=center, font=\scriptsize\sffamily] (n_1) at (0,0.64) {TIAGo Robot Cam};
			\node[label,scale=.5, anchor=south west, rectangle, fill=white, align=center, font=\scriptsize\sffamily] (n_2) at (0.45, 1.) {3D Scene View};
		\end{scope}
  
  \begin{scope}[shift=(image2.north west),x={(image2.north east)},y={(image2.south west)}]
			\node[label, scale=.5, anchor=south west, rectangle, rounded corners=2, inner sep=0.06cm, fill={white!90!black},opacity=.9,text opacity=1, align=center, font=\scriptsize\sffamily] (l_cam2) at (0.42, 0.17) {Cam\,1};
    		\draw[{white!90!black},opacity=.9, very thick] (l_cam2.300) ++(0., -0.002) -- ++(0., 0.05);
    				    
		    \node[label, scale=.5, anchor=south west, rectangle, rounded corners=2, inner sep=0.06cm, fill={white!90!black},opacity=.9,text opacity=1, align=center, font=\scriptsize\sffamily] (l_cam1) at (0.42, 0.38) {Cam\,2};
			\draw[{white!90!black},opacity=.9, very thick] (l_cam1.60) ++(0., 0.002) -- ++(0., -0.04);
    		
    		\node[label, scale=.5, anchor=south west, rectangle, rounded corners=2, inner sep=0.06cm, fill={white!90!black},opacity=.9,text opacity=1, align=center, font=\scriptsize\sffamily] (l_cam4) at (0.84, 0.375) {Cam\,3};
    		\draw[{white!90!black},opacity=.9, very thick] (l_cam4.130) ++(0., 0.002) -- ++(-0., -0.04);
    		
    		\node[label, scale=.5, anchor=south west, rectangle, rounded corners=2, inner sep=0.06cm, fill={white!90!black},opacity=.9,text opacity=1, align=center, font=\scriptsize\sffamily] (l_cam3) at (0.84, 0.17) {Cam\,4};
    		\draw[{white!90!black},opacity=.9, very thick] (l_cam3.225) ++(0., -0.002) -- ++(0., 0.05);
    		
    		 \node[label, scale=.5, anchor=south west, rectangle, rounded corners=2, inner sep=0.06cm, fill={white!90!black},opacity=.9,text opacity=1, align=center, font=\scriptsize\sffamily] (l_camR) at (0.56, 0.38) {Robot\\Cam};
    \draw[{white!90!black},opacity=.9, very thick] (l_camR.347) ++(-0.002, 0.) -- ++(0.03, 0.);
    		
    		\node[label,scale=.5, anchor=north west, rectangle, fill=white, align=center, font=\scriptsize\sffamily] (n_0) at (0,0.005) {External Cam~3};
			\node[label,scale=.5, anchor=north west, rectangle, fill=white, align=center, font=\scriptsize\sffamily] (n_1) at (0,0.44) {HSR Robot Cam};
			\node[label,scale=.5, anchor=south west, rectangle, fill=white, align=center, font=\scriptsize\sffamily] (n_2) at (0.42, 1.) {3D Scene View};
		
    	\end{scope}
    	
   		\begin{scope}[shift=(image3.north west),x={(image3.north east)},y={(image3.south west)}]
			\node[label, scale=.5, anchor=south west, rectangle, rounded corners=2, inner sep=0.06cm, fill={white!90!black},opacity=.9,text opacity=1, align=center, font=\scriptsize\sffamily] (l_cam2) at (0.42, 0.17) {Cam\,1};
    		\draw[{white!90!black},opacity=.9, very thick] (l_cam2.300) ++(0., -0.002) -- ++(0., 0.05);
    				    
		    \node[label, scale=.5, anchor=south west, rectangle, rounded corners=2, inner sep=0.06cm, fill={white!90!black},opacity=.9,text opacity=1, align=center, font=\scriptsize\sffamily] (l_cam1) at (0.42, 0.38) {Cam\,2};
			\draw[{white!90!black},opacity=.9, very thick] (l_cam1.60) ++(0., 0.002) -- ++(0., -0.04);
    		
    		\node[label, scale=.5, anchor=south west, rectangle, rounded corners=2, inner sep=0.06cm, fill={white!90!black},opacity=.9,text opacity=1, align=center, font=\scriptsize\sffamily] (l_cam4) at (0.84, 0.375) {Cam\,3};
    		\draw[{white!90!black},opacity=.9, very thick] (l_cam4.130) ++(0., 0.002) -- ++(-0., -0.04);
    		
    		\node[label, scale=.5, anchor=south west, rectangle, rounded corners=2, inner sep=0.06cm, fill={white!90!black},opacity=.9,text opacity=1, align=center, font=\scriptsize\sffamily] (l_cam3) at (0.84, 0.17) {Cam\,4};
    		\draw[{white!90!black},opacity=.9, very thick] (l_cam3.225) ++(0., -0.002) -- ++(0., 0.05);
    		
    		 \node[label, scale=.5, anchor=south west, rectangle, rounded corners=2, inner sep=0.06cm, fill={white!90!black},opacity=.9,text opacity=1, align=center, font=\scriptsize\sffamily] (l_camR) at (0.53, 0.395) {Robot\\Cam};
    \draw[{white!90!black},opacity=.9, very thick] (l_camR.347) ++(-0.002, 0.) -- ++(0.03, 0.);
    		
    		\node[label,scale=.5, anchor=north west, rectangle, fill=white, align=center, font=\scriptsize\sffamily] (n_0) at (0,0.005) {External Cam~3};
			\node[label,scale=.5, anchor=north west, rectangle, fill=white, align=center, font=\scriptsize\sffamily] (n_1) at (0,0.44) {HSR Robot Cam};
			\node[label,scale=.5, anchor=south west, rectangle, fill=white, align=center, font=\scriptsize\sffamily] (n_2) at (0.42, 1.) {3D Scene View};
			
		\end{scope}
  \end{tikzpicture}
  \vspace{-2.em}
  \caption{Human-aware anticipatory navigation: Visualization of a person emerging from behind an occluding wall for TIAGo (left) and HSR (right) robots. Semantic feedback about tracked persons and their velocity (red arrow) allows the robots to anticipate and foresightedly adjust their navigation paths.}
  \vspace{-.3em}
  \label{fig:human_aware_nav}
\end{figure*}

The experiment is illustrated in \cref{fig:human_aware_nav} for a scenario where a person emerges from behind an occluding wall and crosses the robot's path.
Without feedback (Column~(a)), the robot can react to the person only after they emerge from behind the wall and are visible in its own sensors. Robot and person come dangerously close.
With semantic feedback from the external sensors about tracked persons (Column~(b)), the person with its linear velocity estimate is included in the robot's dynamic obstacle map, where regions they (prospectively) walk through are marked as occupied (dark gray color) even before they appear from behind the occlusion.
The robot adapts its navigation path foresightedly and keeps a safe distance from the person who is crossing its originally planned path.

In Tab.~\ref{tab:nav}, we provide a quantitative evaluation of the minimum safety distance towards the human maintained by the robot during five iterations of the experiment with and without anticipation, respectively.
The experiment was conducted for two robots, HSR (H) and TIAGo (T), and two subjects (S1 and S2). The HSR experiments were executed only by S1. 
The experiments show that the minimal safety distance is significantly higher with anticipation compared to the default robot behavior. Through the semantic human pose feedback, the robots can anticipate a person emerging from behind an occlusion significantly earlier and anticipatorily adjust their navigation path to always maintain a sufficient safety distance of at least \SI{82}{\centi\meter} or \SI{50}{\centi\meter} for HSR and TIAGo, respectively. Without anticipation, the robots come dangerously close to the emerging persons with a worst-case distance of only \SI{11}{\centi\meter} resp. \SI{8}{\centi\meter}.

\subsubsection{Collaborative Furniture Transport}
\label{sec:eval_collab}

\begin{figure*}[t]
  \centering
  \begin{tikzpicture}
  \node[anchor=north west,inner sep=0] (image) at (0,0){\includegraphics[height=2.05cm]{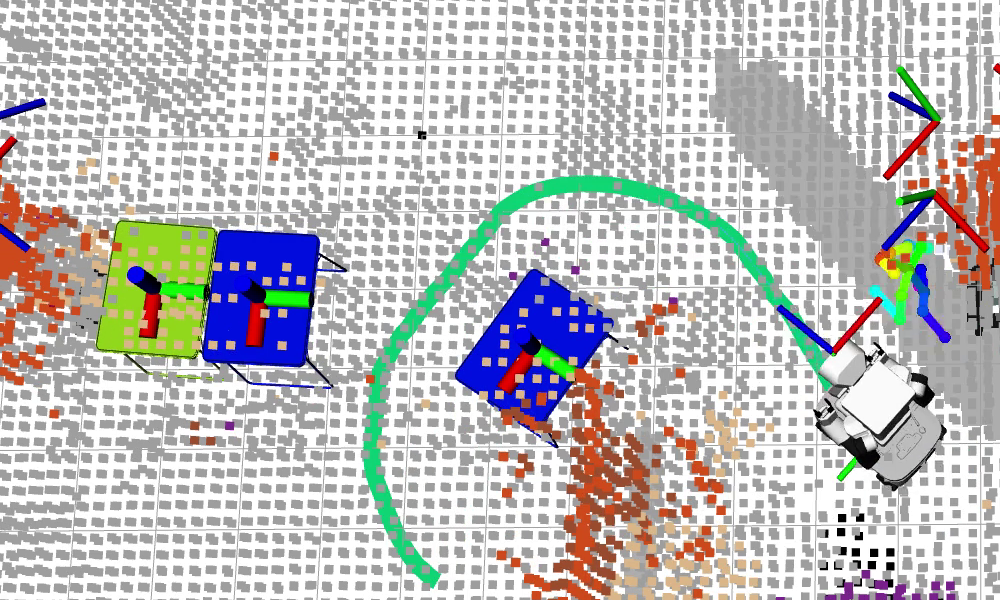}};
  \node[anchor=north west,inner sep=0, xshift=0.2em] (image1) at (image.north east){\includegraphics[height=2.05cm]{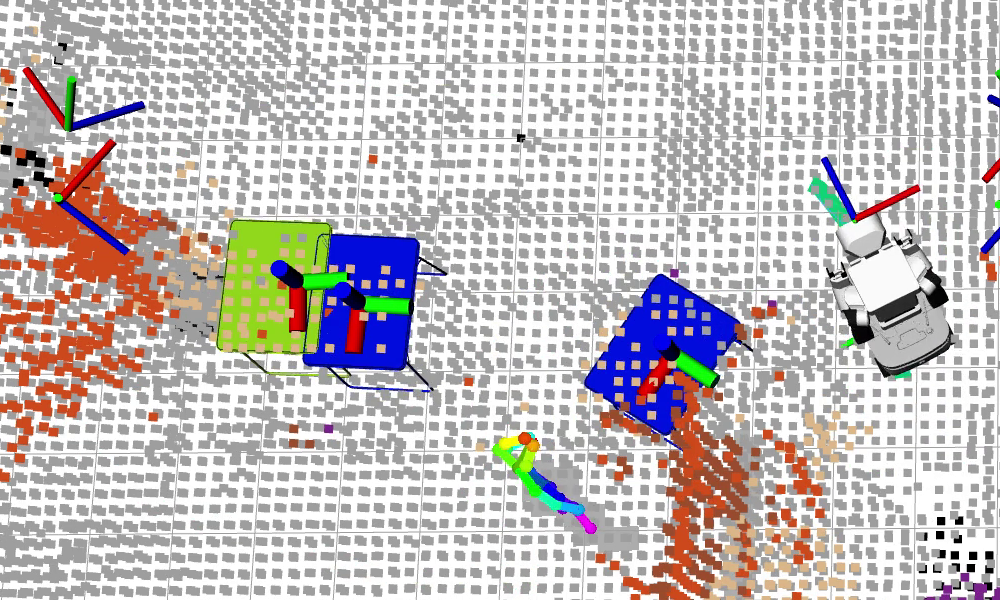}};
  \node[anchor=north west,inner sep=0, xshift=0.2em] (image2) at (image1.north east){\includegraphics[height=2.05cm]{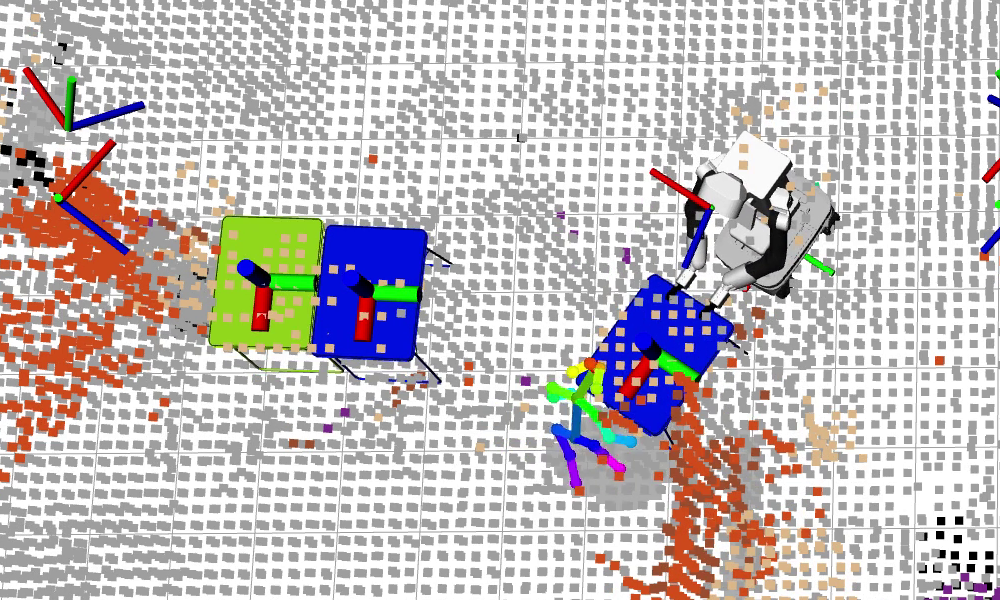}};
  \node[anchor=north west,inner sep=0, xshift=0.2em] (image3) at (image2.north east){\includegraphics[height=2.05cm]{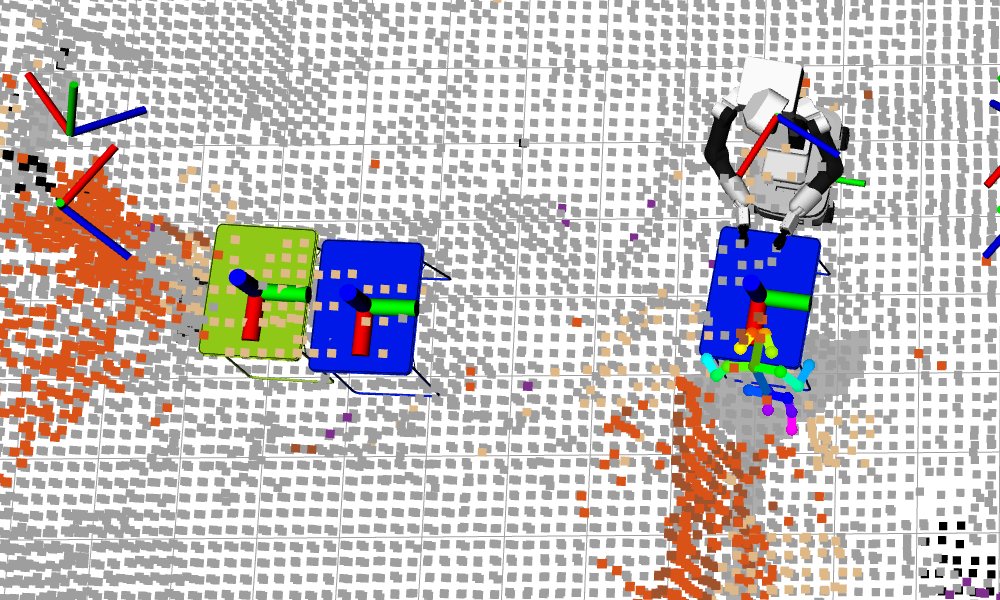}};
  \node[anchor=north west,inner sep=0, xshift=0.2em] (image4) at (image3.north east){\includegraphics[height=2.05cm]{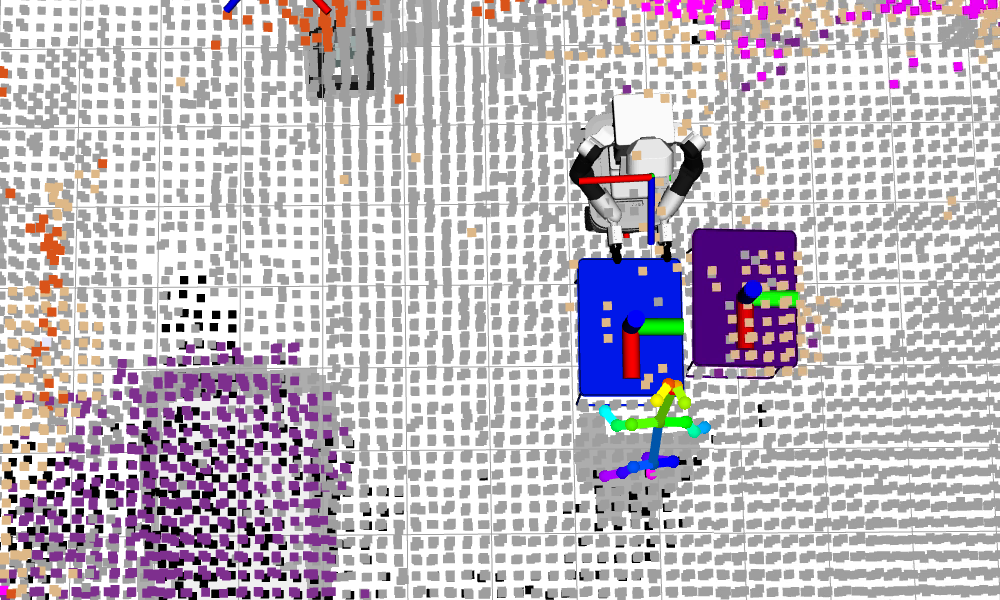}};

  \node[anchor=north west,inner sep=0, yshift=-0.2em] (imagepic)  at (image.south  west){\includegraphics[height=2.05cm]{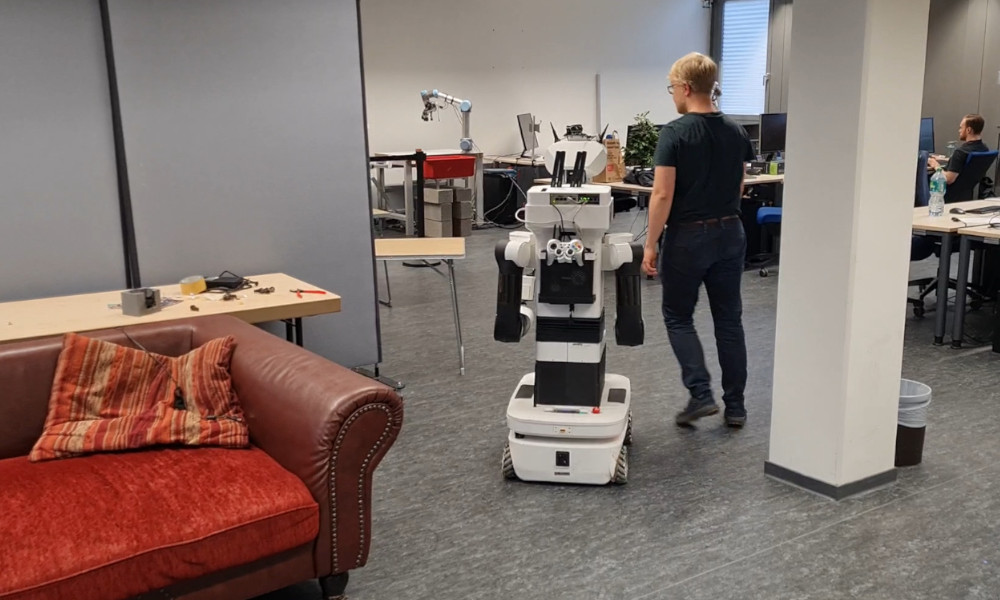}};
  \node[anchor=north west,inner sep=0, yshift=-0.2em] (imagepic1) at (image1.south west){\includegraphics[height=2.05cm]{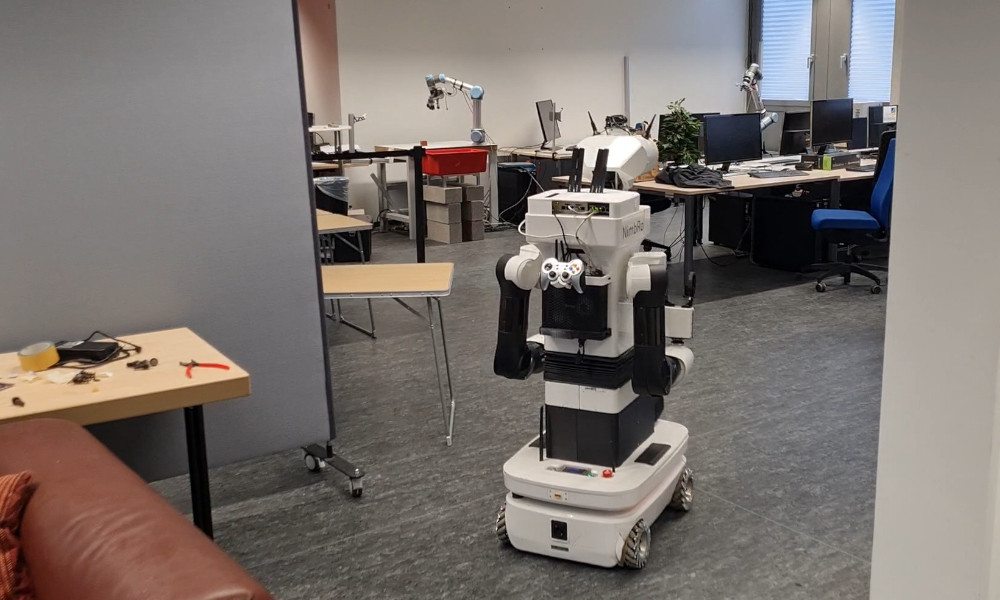}};
  \node[anchor=north west,inner sep=0, yshift=-0.2em] (imagepic2) at (image2.south west){\includegraphics[height=2.05cm]{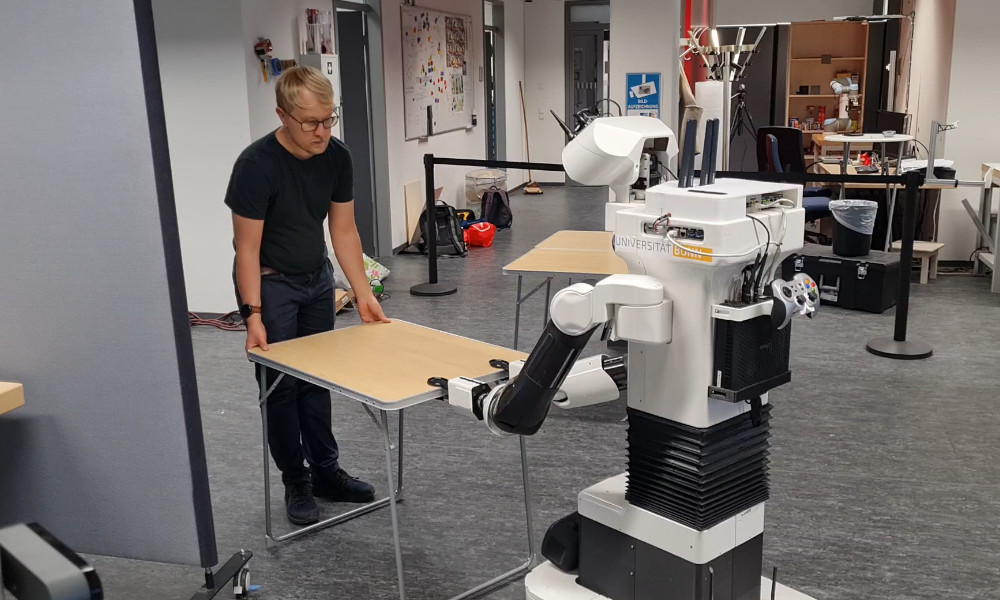}};
  \node[anchor=north west,inner sep=0, yshift=-0.2em] (imagepic3) at (image3.south west){\includegraphics[height=2.05cm]{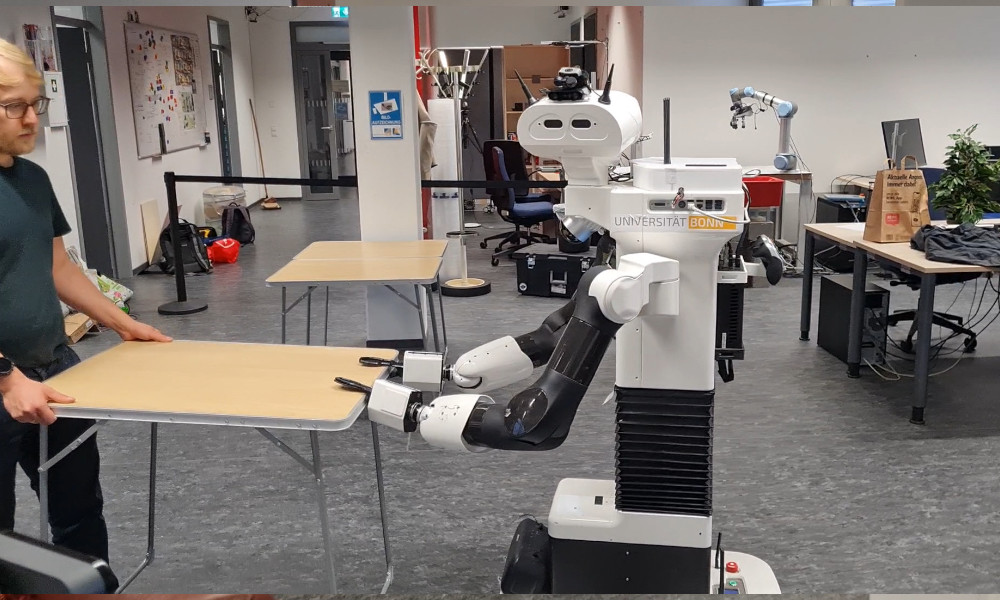}};
  \node[anchor=north west,inner sep=0, yshift=-0.2em] (imagepic4) at (image4.south west){\includegraphics[height=2.05cm]{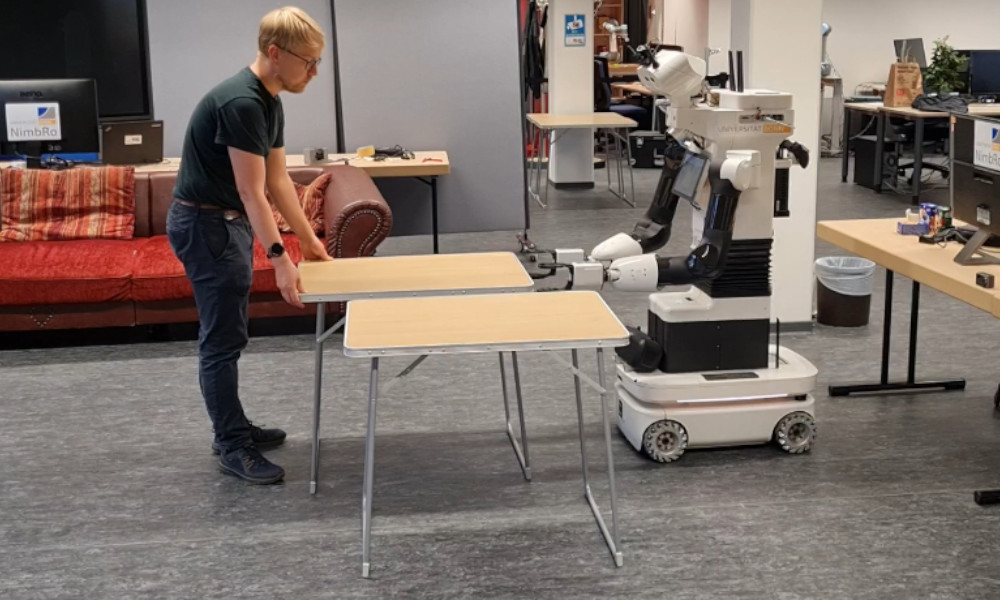}};
  
  \node[inner sep=0.,scale=.9, anchor=north,yshift=-0.3em, rectangle, align=left, font=\scriptsize\sffamily] (n_0) at (imagepic.south) {initial approach};
  \node[inner sep=0.,scale=.9, anchor=north,yshift=-0.3em, rectangle, align=left, font=\scriptsize\sffamily] (n_0) at (imagepic1.south) {adapted approach};
  \node[inner sep=0.,scale=.9, anchor=north,yshift=-0.3em, rectangle, align=left, font=\scriptsize\sffamily] (n_0) at (imagepic2.south) {picking up table};
  \node[inner sep=0.,scale=.9, anchor=north,yshift=-0.3em, rectangle, align=left, font=\scriptsize\sffamily] (n_0) at (imagepic3.south) {collaborative carrying};
  \node[inner sep=0.,scale=.9, anchor=north,yshift=-0.3em, rectangle, align=left, font=\scriptsize\sffamily] (n_0) at (imagepic4.south) {placing table};
  \end{tikzpicture}
  \vspace{-.7em}
  \caption{Collaborative table carrying task. The robot anticipates the person's choice of which table to carry and which side to grasp. The person guides the compliant robot via interaction forces through the table. The robot anticipates the target placement location and takes the lead for precise positioning.}
  \vspace{-1.em}
  \label{fig:collaborative_table_carrying}
\end{figure*}

In a second set of experiments, we evaluate collaborative furniture carrying with the TIAGo robot.
The task consists of collaboratively arranging tables to a user-defined target layout.
The experiment is conducted with and without anticipatory control.
\textit{With active anticipation}, the \smartedge sensor network predicts the human's intention of which table to lift and from which side of the table to assist the human (cf. Sec.~\ref{sec:nav}). Furthermore, the compliant control mode is guided by anticipating the goal location of the table placement (cf. Sec.~\ref{sec:manip}).
\textit{Without anticipation}, the \smartedge sensor network does not provide global context information for anticipation. The robot cannot anticipate the human's intention but waits for human input on a touchscreen interface. The compliant control mode during table carrying is active and translates to simple robot velocity control, without target location anticipation.
Two subjects performed the task with and without anticipation for five repetitions each.
Sample visualizations from the semantic scene model and a camera view are shown in \cref{fig:collaborative_table_carrying}.

\begin{table}[t]
  \centering
  \caption{Average and worst-case person--robot safety dis\-tance.}
  \vspace{-.7em}
  \setlength{\tabcolsep}{4pt}
  \begin{tabular}{lcccccc}
    \toprule
    & \multicolumn{3}{c}{w/o anticipation} & \multicolumn{3}{c}{w/ anticipation} \\
    \cmidrule (lr) {2-4} \cmidrule (lr) {5-7}
    & S1\,--\,H & S1\,--\,T & S2\,--\,T & S1\,--\,H & S1\,--\,T & S2\,--\,T \\
    \midrule
    \textbf{avg.}  &  0.41\,m & 0.23\,m & 0.19\,m & \textbf{0.86\,m} & 0.71\,m & 0.79\,m\\
    \textbf{worst} &  0.11\,m & 0.08\,m & 0.12\,m & \textbf{0.82\,m} & 0.50\,m & 0.61\,m\\
    \bottomrule
    \label{tab:nav}
  \end{tabular}\vspace*{-1ex}\\
	\scriptsize  Results from five runs of the anticipatory navigation experiment\\
	             with HSR (H) and TIAGo (T) robots for two subjects (S1, S2).
\end{table}
\begin{table}[t]
\centering
  \caption{ Collaborative furniture carrying performance.}
  \vspace{-.7em}
  \setlength{\tabcolsep}{4pt}
  \begin{tabular}{ccccc}
    \toprule
    \textbf{Anticipation} &\textbf{Trans. Error (\SI{}{\meter})} & \textbf{Ang. Error (\SI{}{\degree})} &\textbf{Duration (\SI{}{\second})} \\
    \midrule
    \checkmark \qquad S1 & \textbf{0.06} $\pm$ 0.04 & 0.64 $\pm$ 0.57 & \;\,88 $\pm$ 11 \\
    \checkmark \qquad S2 & \textbf{0.06} $\pm$ 0.04 & \textbf{0.36} $\pm$ 0.47 & \;\,\textbf{86} $\pm$ \;\,7 \\
    \midrule
    -- \qquad S1 & 0.13 $\pm$ 0.03 & 6.41 $\pm$ 5.64 & 106 $\pm$ \;\,9 \\
    -- \qquad S2 & 0.29 $\pm$ 0.07 & 14.50 $\pm$ 10.44 & 124 $\pm$ 16 \\
    \bottomrule
  \end{tabular}\vspace*{1ex}
	\scriptsize Experiment with (top) and without anticipation (bottom).\\ The results of each subject are averaged over five trials.
  \label{tab:collaborative_table_carrying_experiment_short}\vspace*{-5mm}
\end{table}
 
We evaluate the task performance in terms of translation and angular error, and task completion time.
The translation error is calculated as the Euclidean distance between the goal pose and the actual placement pose of the robot. Similarly, we calculate the angular error as the yaw angle difference. %
The task achievement time is measured from the start of the task until the table is placed at the goal location.

The results are shown in Tab.~\ref{tab:collaborative_table_carrying_experiment_short}. 
The table carrying task is executed faster with anticipation and the pose error is smaller compared to the task without anticipation. On average, the anticipated collaboration was completed \SI{26}{\second} faster than the user interface-based baseline. Furthermore, the target location is achieved more accurately with anticipation.
\subsection{Collaborative Furniture Rearrangement}

\begin{figure}[t]
  \centering
  \begin{tikzpicture}[scale=0.5]
    \scriptsize
  
    \def\robotpos{(4.,3.7)}
    \def\tablecenter{(5.5, 3.3)}
    \def\tabledirection{(-0.5,0.0)}
    \def\robotdirection{(0.5,0)}
  
    \def\sourcetablepos{(2,2)}
    \def\sourcetableposcenter{(2.5,3)}
    \def\sourcetabledirection{(0,0.5)}
  
    \def\humanpos{(7.1,2.8)}
    \def\humandirection{(-1.0,0)}

    \def\endeffectorcenter{(-1.,0.2)}
  
    \def\targettablepos{(9,0.75)}
    \def\targettableposwithrobotoffset{(9,1.5)}

    \def\targetrobotpos{(9,2.25)}
  
    \def\chairpickuppos{(14,0)}

  	\draw[thick, gray] (-1.5,5.5) rectangle ++ (17.2,-10.6);
  	
    \draw[thick, dotted] (-1,1) rectangle ++ (5,4);
    \node[anchor=north] at (1,1) {Table Pickup $\mathcal{A}_\text{pick}^\text{table}$};

    \draw[thick, dotted] (7,2) rectangle ++ (5,-6);
    \node[anchor=north] at (8.5,-4.0) {Target Layout $\mathcal{L}$};

    \draw[thick, dotted] (13,2) rectangle ++ (2,-6);
    \node[anchor=north] at (13.5,-4.0) {Chair Pickup $\mathcal{A}_\text{pick}^\text{chair}$};

    \draw[thick, fill=blue!20, rotate around={-3:\sourcetablepos}] \sourcetablepos rectangle ++ (1,2);
    \draw[thick, ->] \sourcetableposcenter -- ++ (0.05,0.5); %
    \draw[thick, fill=black] \sourcetableposcenter circle (0.05);

    \draw[thick, fill=blue!20] \sourcetablepos ++ (-1.4,0) rectangle ++ (1,2);
    \draw[thick, fill=black] \sourcetableposcenter ++ (-1.4,0) circle (0.05);
    \draw[thick, ->] \sourcetableposcenter ++ (-1.4,0) -- ++ \sourcetabledirection;
  
    \draw[thick, fill=blue!20] \sourcetablepos ++ (-2.6,0) rectangle ++ (1,2);
    \draw[thick, fill=black] \sourcetableposcenter ++ (-2.6,0) circle (0.05);
    \draw[thick, ->] \sourcetableposcenter ++ (-2.6,0) -- ++ \sourcetabledirection;

    \draw[thick, fill=cyan!20] \chairpickuppos ++ (-.5,-.5) rectangle ++ (1,1);
    \draw[thick, ->] \chairpickuppos -- ++ (0.5,0.0);
    \draw[thick, fill=black] \chairpickuppos circle (0.05);
  
    \draw[thick, fill=cyan!20] \chairpickuppos ++ (0,-2.2) ++ (-.5,-.5) rectangle ++ (1,1);
    \draw[thick, ->] \chairpickuppos ++ (0,-2.2) -- ++ (0.5,0.0);
    \draw[thick, fill=black] \chairpickuppos ++ (0,-2.2) circle (0.05);

    \draw[thick, fill=red!10, dashed] \targetrobotpos circle (0.5);
    \draw[thick, fill=black] \targetrobotpos circle (0.05) node [anchor=east] {$\mathbf{p}_{\text{goal}}$};
    \draw[thick,->, dashed] \targetrobotpos -- ++ (0., -0.5);

    \draw[thick, dashed, fill=green!20] \targettablepos ++ (-.5,-1.0) rectangle ++ (1,2);
    \draw[thick, dashed, ->] \targettablepos -- ++ (0,0.5);
    \draw[thick, fill=black] \targettablepos circle (0.05);

    \draw[thick, dashed, pattern color=green!20, pattern=crosshatch] \targettablepos ++ (1.2,0) ++ (-.5,-1) rectangle ++ (1,2);
    \draw[thick, dashed, ->] \targettablepos ++ (1.2,0) -- ++ (0,0.5);
    \draw[thick, fill=black] \targettablepos ++ (1.2,0) circle (0.05);

    \draw[thick, dashed, pattern color=lime!20, pattern=crosshatch] \targettablepos ++ (1.85,0) ++ (-.5,-.5) rectangle ++ (1,1);
    \draw[thick, dashed, ->] \targettablepos ++ (1.85,0) -- ++ (-0.5,0.0);
    \draw[thick, fill=black] \targettablepos ++ (1.85,0) circle (0.05);

    \def\verticaloffset{-3.4}

    \draw[thick, dashed, pattern color=green!20, pattern=crosshatch] \targettablepos ++ (1.2,\verticaloffset) ++ (-.5,-1) rectangle ++ (1,2);
    \draw[thick, dashed, ->] \targettablepos ++ (1.2,\verticaloffset) -- ++ (0,0.5);
    \draw[thick, fill=black] \targettablepos ++ (1.2,\verticaloffset) circle (0.05);

    \draw[thick, dashed, pattern color=lime!20, pattern=crosshatch]  \targettablepos ++ (1.85,\verticaloffset) ++ (-.5,-.5) rectangle ++ (1,1);
    \draw[thick, dashed, ->] \targettablepos ++ (1.85,\verticaloffset) -- ++ (-0.5,0.0);
    \draw[thick, fill=black] \targettablepos ++ (1.85,\verticaloffset) circle (0.05);
  
    \draw[thick, dashed, pattern color=green!20, pattern=crosshatch] \targettablepos ++ (0,\verticaloffset) ++ (-.5,-1) rectangle ++ (1,2);
    \draw[thick, dashed, ->] \targettablepos ++ (0,\verticaloffset) -- ++ (0,0.5);
    \draw[thick, fill=black] \targettablepos ++ (0,\verticaloffset) circle (0.05);

    \draw[thick, fill=yellow!20] \humanpos circle (0.5);
    \draw[thick, fill=black] \humanpos circle (0.05);
    \node[anchor=north, yshift=1.5, xshift=-2] at \humanpos {$\mathbf{x}_{\text{p}}$};

    \draw[thick, fill=red!20] \robotpos circle (0.5);
    \draw[thick,->, rotate around={-15:\robotpos}] \robotpos -- ++\robotdirection;
    \draw[thick, fill=black] \robotpos circle (0.05);
    \node[anchor=east, xshift=5., yshift=-3.] at \robotpos {$\mathbf{p}_{\text{r}}$};

    \draw[thick, fill=orange!20, rotate around={-15:\tablecenter}] \tablecenter ++ (-1,-0.5) rectangle ++ (2,1);
    \draw[thick,->, rotate around={-15:\tablecenter}] \tablecenter -- ++ \tabledirection;
    \draw[thick, fill=black] \tablecenter circle (0.05);

  \end{tikzpicture}
  \vspace{-1.em}
\caption{Collaborative furniture arrangement: Tables and chairs are carried collaboratively with the human from $\mathcal{A}_\text{pick}^\text{table}$ resp. $\mathcal{A}_\text{pick}^\text{chair}$ to the user-defined target layout $\mathcal{L}$. Human, robot, and object poses are estimated by the \smartedge sensor network and used to anticipate and navigate to the pickup pose (Alg.~\ref{alg:nav},~\ref{alg:approach}). During collaborative carrying, the robot anticipates the target table pose based on $\mathcal{L}$ (Alg.~\ref{alg:followhand}). Chairs can be moved by the robot alone.}
  \label{fig:collaborative_furniture_arrangement}
  \vspace{-0.5cm}
\end{figure}

\begin{figure*}[t]
  \centering
  \begin{tikzpicture}
  \node[anchor=north west,inner sep=0] (image) at (0,0){\includegraphics[height=3.0cm]{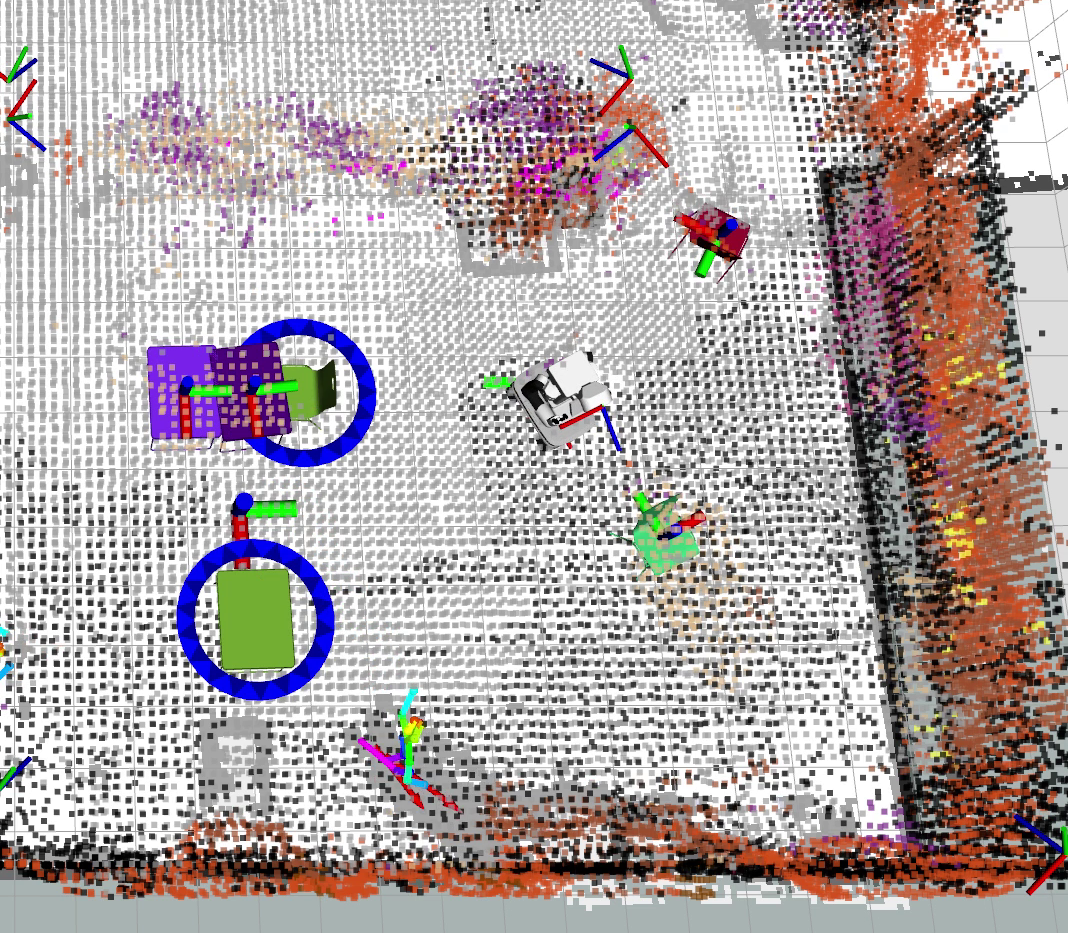}};
  \node[anchor=north west,inner sep=0, xshift=0.2em] (image1) at (image.north east){\includegraphics[height=3.0cm]{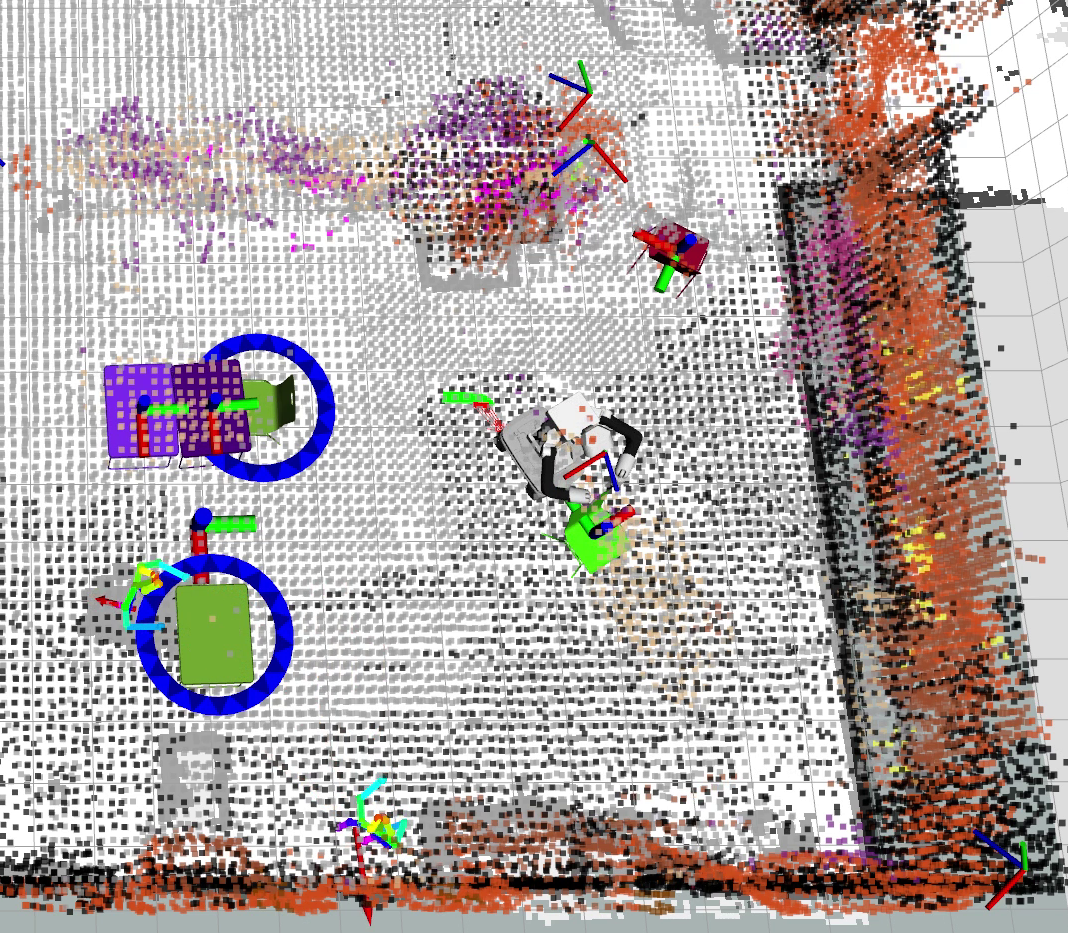}};
  \node[anchor=north west,inner sep=0, xshift=0.2em] (image2) at (image1.north east){\includegraphics[height=3.0cm]{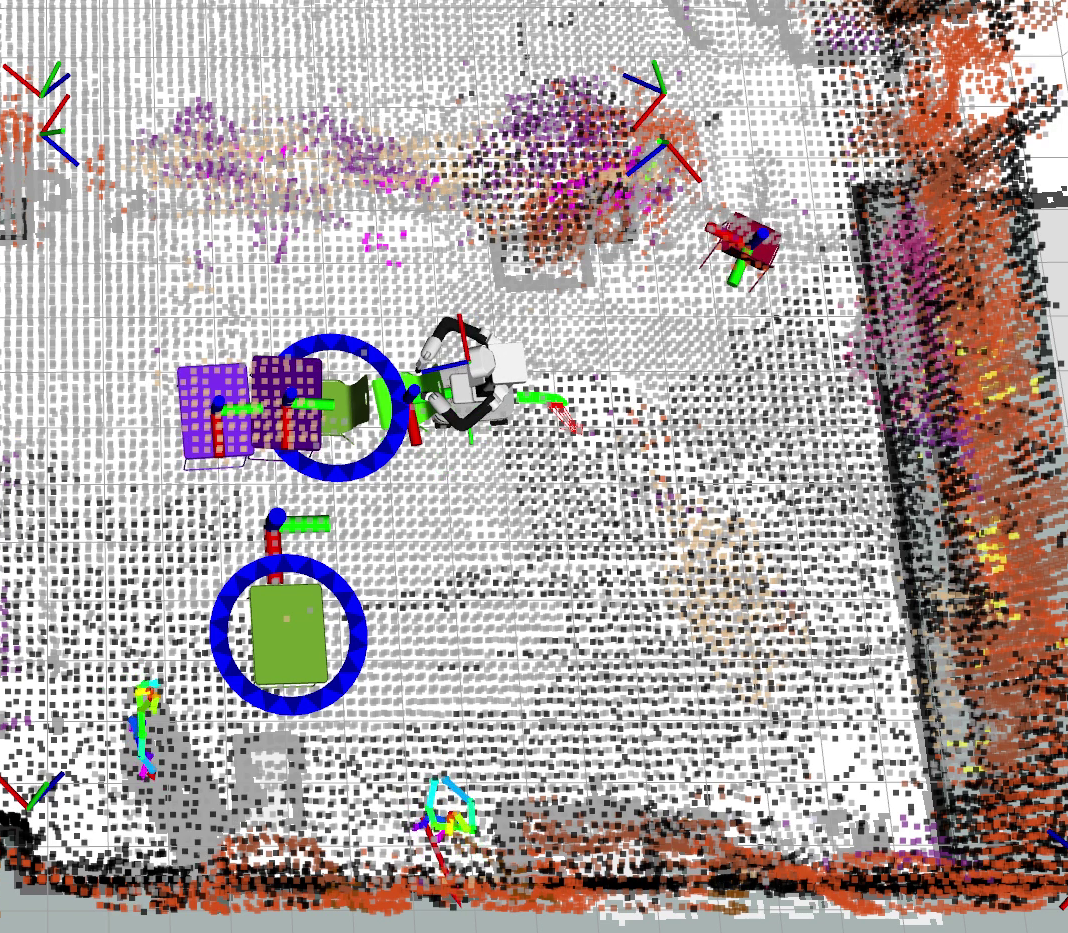}};
  \node[anchor=north west,inner sep=0, xshift=0.2em] (image3) at (image2.north east){\includegraphics[height=3.0cm]{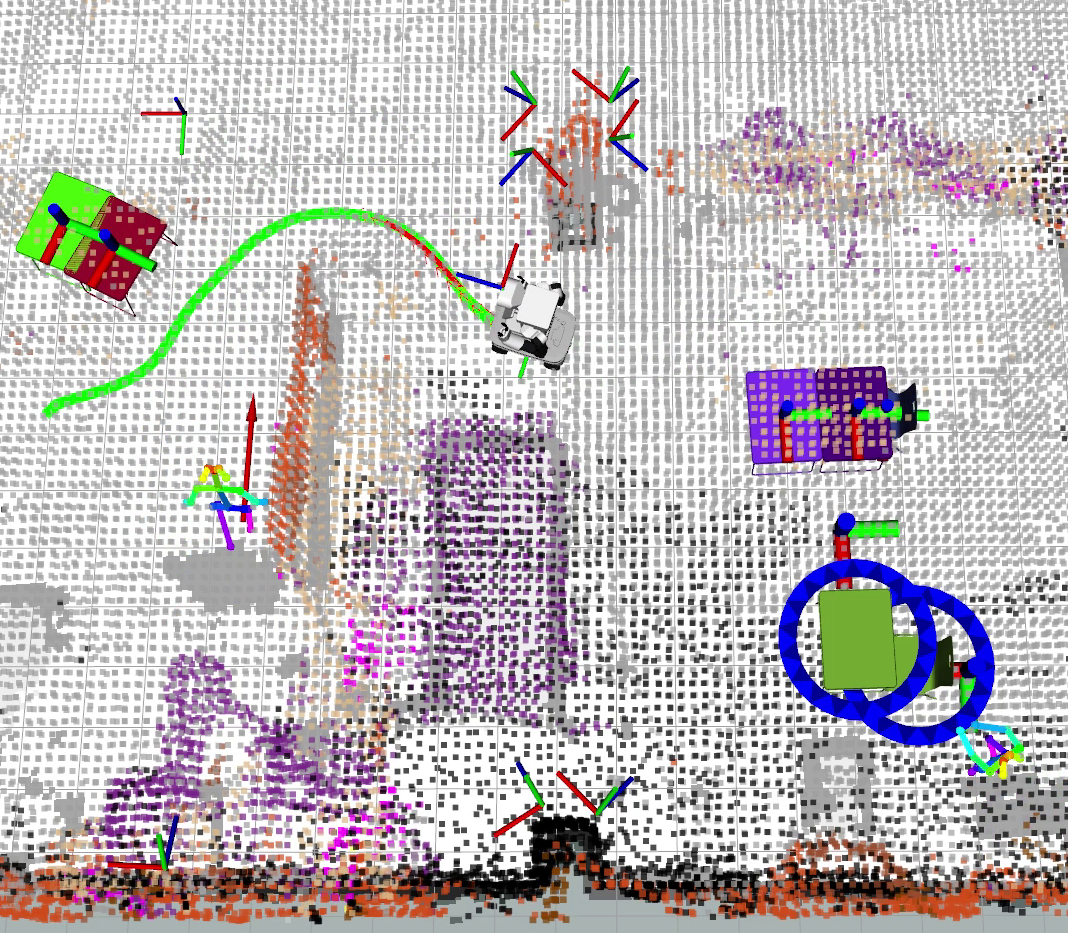}};
  \node[anchor=north west,inner sep=0, xshift=0.2em] (image4) at (image3.north east){\includegraphics[height=3.0cm]{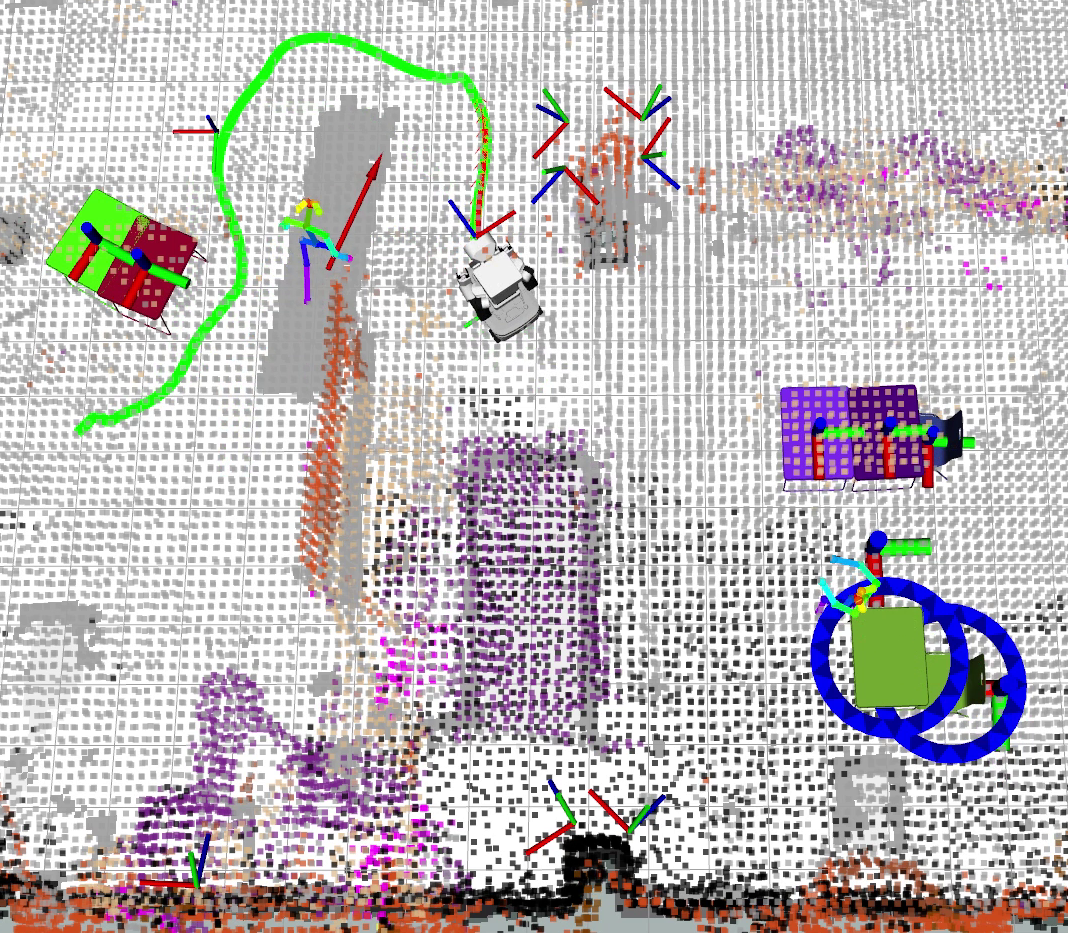}};

  \node[anchor=north west,inner sep=0, yshift=-0.2em] (imagepic)  at (image.south  west){\includegraphics[height=3.0cm]{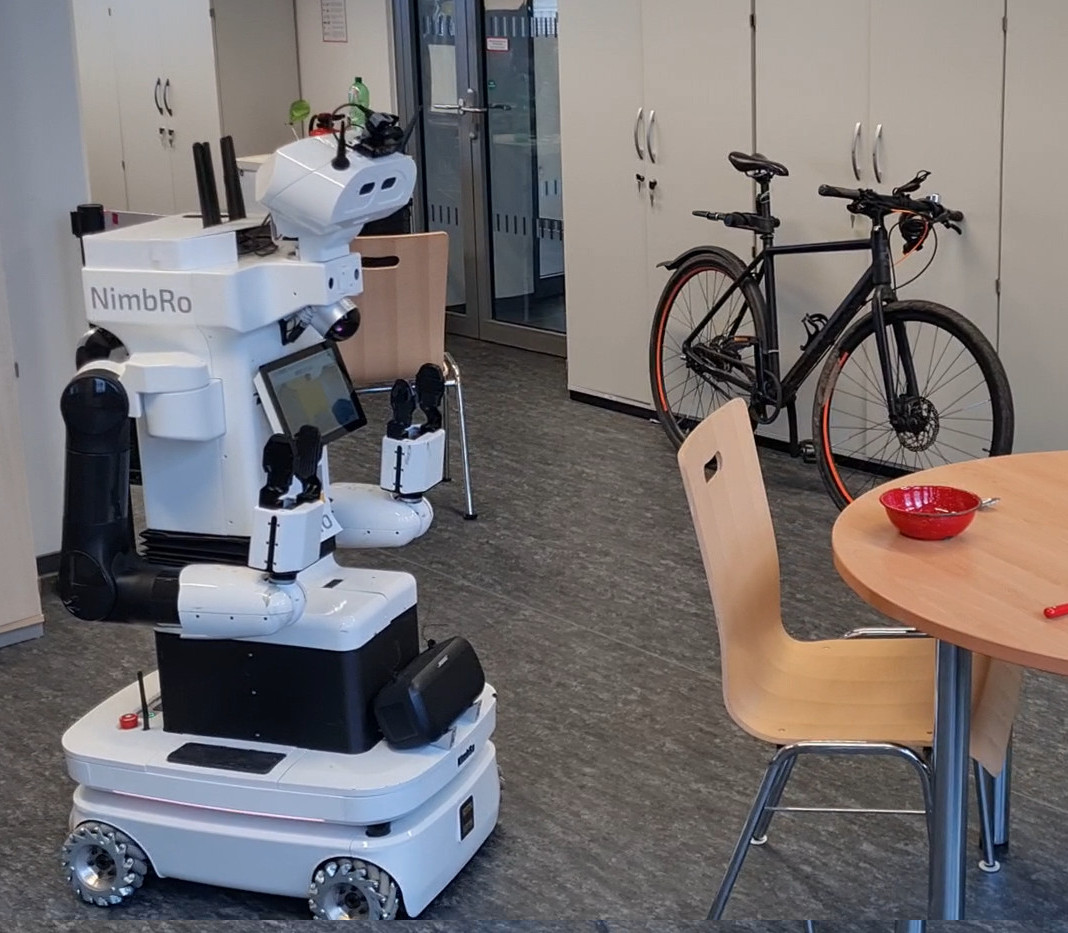}};
  \node[anchor=north west,inner sep=0, yshift=-0.2em] (imagepic1) at (image1.south west){\includegraphics[height=3.0cm]{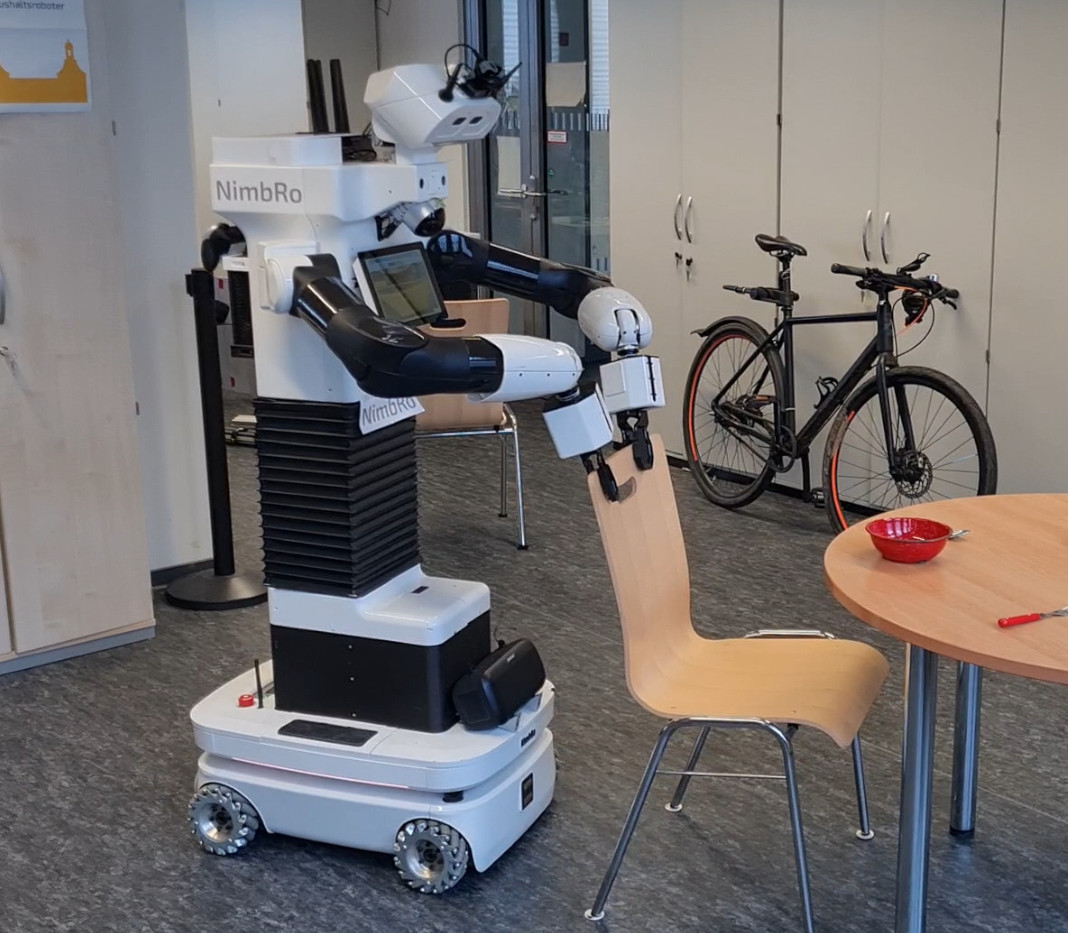}};
  \node[anchor=north west,inner sep=0, yshift=-0.2em] (imagepic2) at (image2.south west){\includegraphics[height=3.0cm]{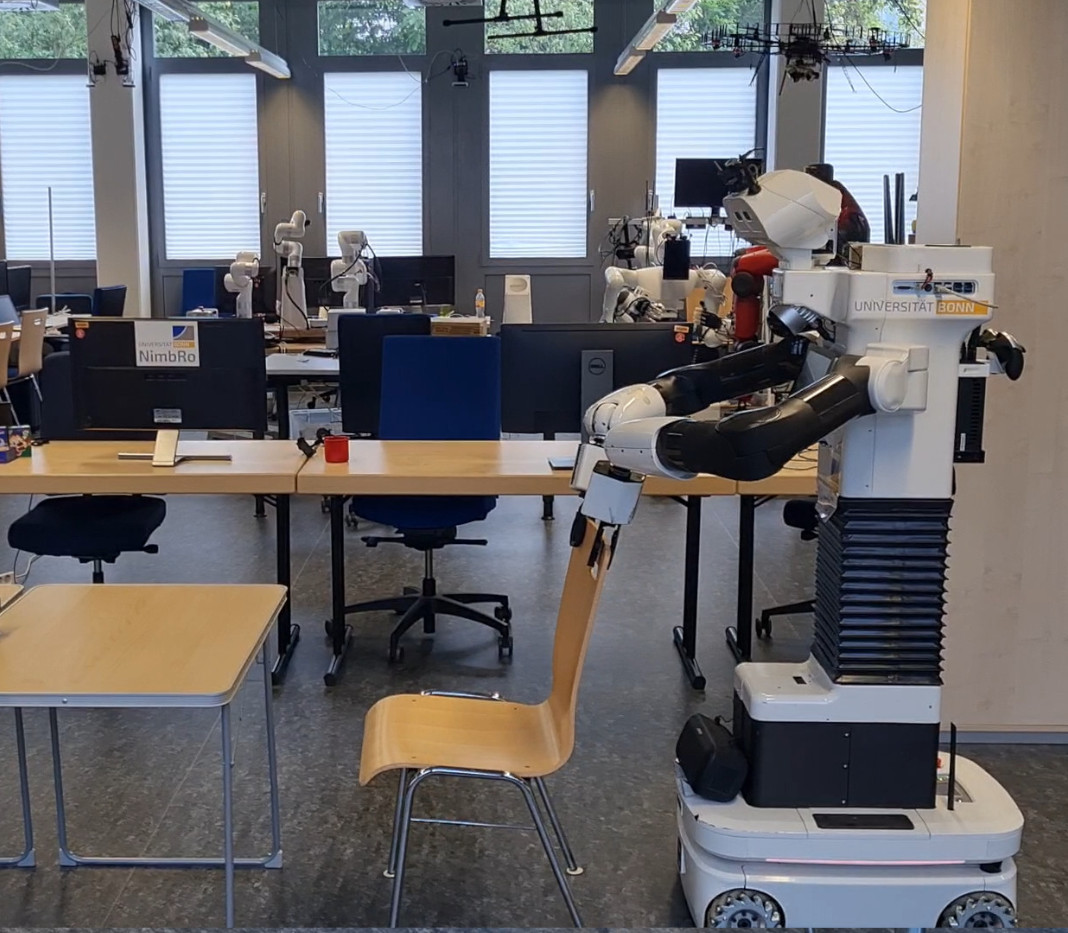}};
  \node[anchor=north west,inner sep=0, yshift=-0.2em] (imagepic3) at (image3.south west){\includegraphics[height=3.0cm]{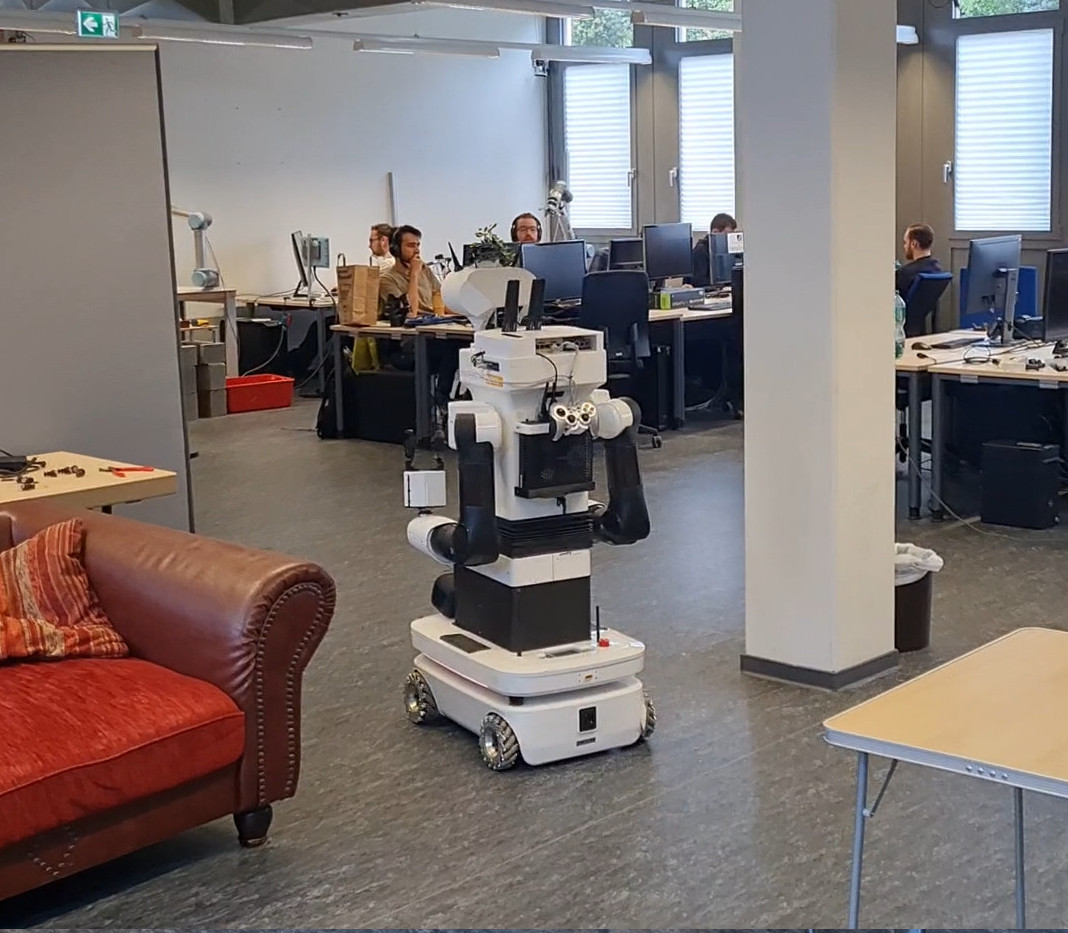}};
  \node[anchor=north west,inner sep=0, yshift=-0.2em] (imagepic4) at (image4.south west){\includegraphics[height=3.0cm]{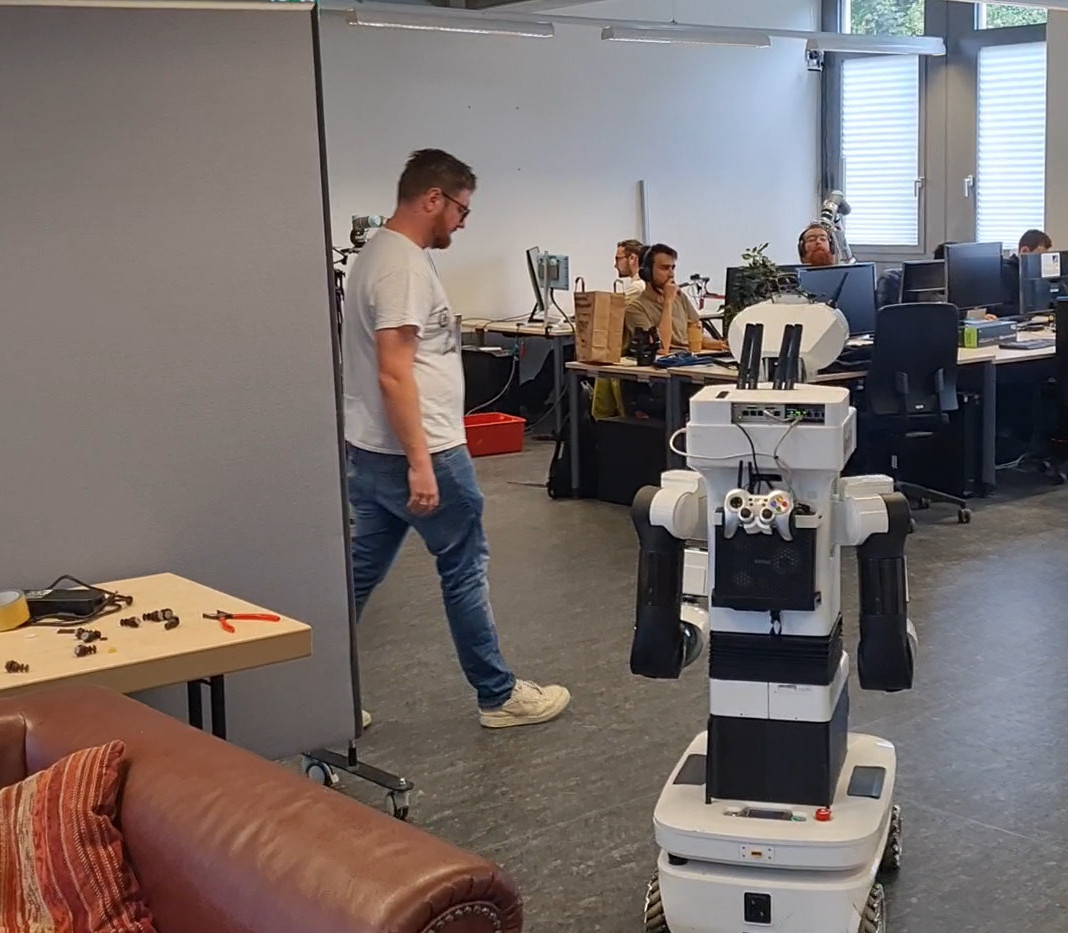}};
  
  \node[inner sep=0.,scale=.9, anchor=north,yshift=-0.3em, rectangle, align=left, font=\scriptsize\sffamily] (n_0) at (imagepic.south) {approaching chair};
  \node[inner sep=0.,scale=.9, anchor=north,yshift=-0.3em, rectangle, align=left, font=\scriptsize\sffamily] (n_0) at (imagepic1.south) {grasping chair};
  \node[inner sep=0.,scale=.9, anchor=north,yshift=-0.3em, rectangle, align=left, font=\scriptsize\sffamily] (n_0) at (imagepic2.south) {pushing chair};
  \node[inner sep=0.,scale=.9, anchor=north,yshift=-0.3em, rectangle, align=left, font=\scriptsize\sffamily] (n_0) at (imagepic3.south) {approaching next table};
  \node[inner sep=0.,scale=.9, anchor=north,yshift=-0.3em, rectangle, align=left, font=\scriptsize\sffamily] (n_0) at (imagepic4.south) {anticipatory navigation};
  \end{tikzpicture}
  \vspace{-.7em}
  \caption{Visualizations of the collaborative furniture arrangement. Human, robot, and object poses are tracked by the smart edge sensor network. Green markers with blue circles denote the anticipated target pose for the next chair resp. table.
  The robot autonomously approaches the chair, grasps and pushes it to the goal pose given by the target layout.
  It then approaches the next table, preemptively adapting its navigation path to the person appearing from behind the occluding wall.}
  \vspace{-0.3cm}
  \label{fig:final_demo}
\end{figure*}

Finally, we show the integration of our anticipation approaches to demonstrate their potential in a collaborative task of arranging furniture to a target room layout as depicted in \cref{fig:collaborative_furniture_arrangement}.
This demonstration combines anticipatory human-aware navigation and collaborative furniture arrangement, including collaboratively carrying tables and moving chairs.
During the 11-minute continuous operation, four tables were carried and two chairs were moved to the previously placed tables. 
For each table, it was anticipated which one should be carried and from which side, based on the \smartedge sensor semantic scene model.
In addition, the robot anticipated the desired placement location based on the room layout while compliantly collaborating with the human to carry the table.
The expected target pose was automatically updated based on the location in the furniture plan the person was approaching.
The chairs to align were also identified by the \smartedge sensors.
When navigating back to the pickup location, the robot received semantic feedback about the predicted movement of persons, also behind occlusions, and anticipatorily adapted its navigation path to avoid collisions and keep a sufficient safety distance.
\cref{fig:final_demo} depicts excerpts of the demonstration, which was conducted with two different subjects.
Please also refer to the accompanying video.

\section{Conclusions}

In this paper, we presented approaches to incorporate allocentric semantic context information from \smartedge sensor network observations to anticipate human behavior on two levels:
(1) in the context of human-aware navigation to improve safety, by projecting future human trajectories into the planning map of a mobile robot, and
(2) in the context of collaborative mobile manipulation for improving efficiency, by anticipating intentions to work towards a desired goal.

Both approaches were evaluated in real-world experiments and compared against non-anticipatory baseline approaches utilizing a graphical user interface for human-robot interaction. Our approach demonstrates safer human-aware navigation and improved efficiency for human-robot collaboration with a mobile manipulation robot.
We show that the robot anticipates persons emerging from behind occlusions and preemptively adjusts its navigation path to maintain a safe distance by incorporating semantic feedback of human pose observations from external sensors.

An integrated demonstration shows our approach's potential for collaborative human-robot interaction, achieving the complex task of setting a room layout with tables and chairs.

Directions for future work include implementing anticipatory human-aware navigation also on a higher planning level, instead of using the local obstacle cost map, taking long-term goals and intents of the persons into account.

\printbibliography

\end{document}